\documentclass[review]{elsarticle}
\graphicspath{ {./figures/} }
\usepackage{hyperref}
\usepackage{float}
\usepackage{verbatim} 
\usepackage{apalike}
\restylefloat{figure}
\restylefloat{table}
\usepackage{algorithm}
\usepackage{algorithmic}
\usepackage{xcolor}
\usepackage{tabularx} 
\usepackage{amsmath}
\usepackage{adjustbox}

\usepackage{listings}
\definecolor{codegreen}{rgb}{0,0.6,0}
\definecolor{codegray}{rgb}{0.5,0.5,0.5}
\definecolor{codepurple}{rgb}{0.58,0,0.82}
\definecolor{backcolour}{rgb}{0.95,0.95,0.92}
\lstdefinestyle{mystyle}{
    backgroundcolor=\color{backcolour},   
    commentstyle=\color{codegreen},
    keywordstyle=\color{magenta},
    numberstyle=\tiny\color{codegray},
    stringstyle=\color{codepurple},
    basicstyle=\ttfamily\footnotesize,
    breakatwhitespace=false,         
    breaklines=true,                 
    captionpos=b,                    
    keepspaces=true,                 
    numbers=left,                    
    numbersep=5pt,                  
    showspaces=false,                
    showstringspaces=false,
    showtabs=false,                  
    tabsize=2
}
\journal{arXiv}

\biboptions{authoryear}

\begin{document}
\begin{frontmatter}

\begin{titlepage}
\begin{center}
\vspace*{1cm}





\textbf{Dermacen Analytica: A Novel Methodology Integrating Multi-Modal Large Language Models with Machine Learning in tele-dermatology}

\vspace{1.5cm}

Dimitrios P. Panagoulias$^{a}$ (panagoulias\_d@unipi.gr), 
Evridiki Tsoureli-Nikita$^{b}$ (evinikita@gmail.com),
Maria Virvou$^{a}$ (mvirvou@unipi.gr), George A. Tsihrintzis$^{a}$ (geoatsi@unipi.gr) \\

\hspace{10pt}

\begin{flushleft}
\small  
$^a$ Department of Informatics, University of Piraeus 185 34, Greece \\

$^b$Athens University Medical School, Athens 115 27, Greece\\

\vspace{1cm}
\textbf{Corresponding Author:} \\
George A. Tsihrintzis \\
Department of Informatics, University of Piraeus 185 34, Greece \\
Tel: (+30) 697 2882168 \\
Email: geoatsi@unipi.gr

\end{flushleft}        
\end{center}
\end{titlepage}
\title{Dermacen Analytica: A Novel Methodology Integrating Multi-Modal Large Language Models with Machine Learning in tele-dermatology}


\author[label1]{Dimitrios P. Panagoulias}
\ead{panagoulias\_d@unipi.gr}
\author[label2]{Evridiki Tsoureli-Nikita}
\ead{evinikita@gmail.com}
\author[label1]{Maria Virvou}
\ead{mvirvou@unipi.gr}

\author[label1, cor1]{George A. Tsihrintzis}
\ead{geoatsi@unipi.gr}

\cortext[cor1]{Corresponding author.}
\address[label1]{Department of Informatics, University of Piraeus, Piraeus 185 34, Greece}
\address[label2]{Athens University Medical School, Athens, Greece}

\begin{abstract}

The rise of Artificial Intelligence creates great promise in the field of medical discovery, diagnostics and patient management. However, the vast complexity of all medical domains require a more complex approach that combines machine learning algorithms, classifiers, segmentation algorithms and, lately, large language models. In this paper, we describe, implement and assess an Artificial Intelligence-empowered system and methodology aimed at assisting the diagnosis process of skin lesions and other skin conditions within the field of dermatology that aims to holistically address the diagnostic process in this domain. The workflow integrates large language, transformer-based vision models and sophisticated machine learning tools. This holistic approach achieves a nuanced interpretation of dermatological conditions that simulates and facilitates a dermatologist's workflow. We assess our proposed methodology through a thorough cross-model validation technique embedded in an evaluation pipeline that utilizes publicly available medical case studies of skin conditions and relevant images. To quantitatively score the system performance, advanced machine learning and natural language processing tools are employed which focus on similarity comparison and natural language inference. Additionally, we incorporate a human expert evaluation process based on a structured checklist to further validate our results. We implemented the proposed methodology in a system which achieved approximate (weighted) scores of 0.87 for both contextual understanding and diagnostic accuracy, demonstrating the efficacy of our approach in enhancing dermatological analysis. The proposed methodology is expected to prove useful in the development of next-generation tele-dermatology applications, enhancing remote consultation capabilities and access to care, especially in underserved areas.
\end{abstract}

\end{frontmatter}

 



\section{Introduction}
\label{introduction}

The rise of Artificial Intelligence  (AI) and its transformative effect on many fields and domains provides significant opportunities for the revamping and re-engineering of the diagnostic process. However, in order to navigate through the complexity of the medical field effectively, a multifaceted approach is essential that integrates several AI technologies including explainability \citep{PANAGOULIAS2024123375}, trustworthiness and holistic evaluation properties \citep{virvouartificial}. Machine learning algorithms, for instance, are instrumental in identifying patterns and insights from large datasets, enabling the prediction of disease progression and response to treatments \citep{ahsan2022machine}. Classifiers play a critical role in categorizing data, such as distinguishing between different types of tumors or identifying patients at high risk for certain conditions \citep{johnson2021precision}. Segmentation algorithms are crucial in imaging analysis, helping to isolate and analyze specific areas within medical images for more accurate diagnoses \citep{wang2022medical}.

In dermatology, while non-invasive methods and imaging techniques offer valuable preliminary assessments, the definitive diagnosis of malignancy relies on laboratory tests and biopsies. This approach ensures accuracy in diagnosis, aids in the planning of appropriate treatments, and helps to avoid the pitfalls of over-diagnosis. The ongoing development of diagnostic technologies aims to refine the accuracy of non-invasive assessments, but the role of histopathological examination remains central to the confirmatory diagnosis of skin cancers and other dermatological conditions.

In this paper, we propose and assess  ``Dermacen Analytica", an assistive workflow designed to support the evaluation of skin lesions and skin conditions and to enhance both the timeliness and efficiency of dermatological analyses. The proposed workflow integrates cutting-edge technologies, including Large Language Models (LLMs), transformer-based vision models, and advanced machine learning tools, to establish a comprehensive solution that addresses the current limitations in tele-dermatology.

More specifically, the key components of our proposed workflow are the (1) Integration of Large Language Models and Transformers with Vision, (2) Advanced Feature Extraction and Segmentation, (3) the Pragmatic Criteria for Skin Condition Evaluation and (4) the Streamlining of the process. 
As per  (1), by leveraging LLMs and vision transformers, our workflow introduces an efficient approach to understanding and interpreting complex dermatological conditions. These models perform well in analyzing visual data, allowing for a nuanced assessment of skin lesions that surpasses traditional analysis methods. As per (2), utilizing state-of-the-art machine learning (ML) tools, the proposed workflow includes a robust feature extraction system that identifies and quantifies key characteristics of skin lesions, such as shape, size, color, and texture. Segmentation algorithms further refine the analysis by isolating lesions from the surrounding skin, enabling precise evaluations. As per (3), our methodology incorporates pragmatic criteria tailored specifically for skin condition evaluation. This includes the consideration of dermatological standards and guidelines, Lastly (4), our workflow demonstrates the potential to serve as the foundation for next-generation tele-dermatology applications. By facilitating remote consultations and evaluations, it expands access to dermatological care, particularly in underserved regions. Its efficiency and accuracy improvements also promise to reduce wait times and enhance patient outcomes.

More specifically, the paper is structured as follows: Section~2 focuses on reviewing relevant previous studies and related published works. In Section~3, Dermacen Analytica is outlined, alongside the evaluation methodology used to validate its use.  In Section~4, we present implementation use cases of our system and the evaluation results. Lastly in Section~5, we summarize the findings and highlights of our research and discuss future related endeavours.

\section{Related work}

Machine learning (ML), particularly its sub-field of deep learning (DL), has become increasingly popular in dermatology due to its ability to learn from large datasets of skin images \citep{chan2020machine,du2020review, hogarty2020artificial}. These algorithms, especially convolutional neural networks (CNNs), have shown proficiency in identifying patterns and features in images that may be subtle or invisible to the human eye \citep{liopyris2022artificial}. This capability is particularly useful in dermatology, where the visual inspection of skin lesions is a primary diagnostic method. The simplest form of ML-based dermatological image assessment involves binary classification, where an ML model is trained to differentiate between two classes of dermatological conditions, namely benign vs. malignant lesions. This is crucial for early detection of skin cancers like melanoma, potentially saving lives through early intervention. Beyond binary classification, ML models can be trained to identify various skin conditions from a single image. This involves categorizing lesions into multiple classes, such as melanoma, basal cell carcinoma, squamous cell carcinoma, and benign nevi among others. In general, multi-class classification systems require more complex models and larger, well-annotated datasets for training.

\subsection{The Dermatologist's Workflow}
The process of diagnosing a skin lesion or condition by dermatologists involves several key steps and requires a thorough consideration and evaluation approach \citep{marwaha2019comparative,gachon2005first}. This process can vary depending on the specific circumstances, but it generally includes the following stages:
\begin{itemize}
    \item \textit{Patient History}: The dermatologist begins by gathering a comprehensive patient history. This involves concerns and symptoms related to conditions, personal or family history of skin diseases, and any other health conditions or medications that could be relevant.
    \item \textit{Visual Examination}: A thorough visual examination of the skin is conducted. Dermatologists are trained to notice subtle differences in skin lesions and can often make a preliminary diagnosis based on their appearance, shape, color, size, texture, and location. They may also examine other areas of the skin to check for additional lesions or signs that could provide diagnostic clues.
    \item \textit{Dermatoscopy}: For suspicious skin lesions, especially those that could be malignant (such as melanoma), a dermatoscope is utilised, which is a specialized magnifying tool with a light source. Dermatoscopy allows the dermatologist to examine the lesion in greater detail, including patterns and colors not visible to the naked eye. The most common rule used for evaluating a skin lesion, is the following ABCDE rule \citep{RING2021635, duarte2021clinical}, which, however, is considered as a guideline and not as a diagnosis tool. Any skin lesion or mole that exhibits one or more of these characteristics usually prompts  further investigation and, if necessary, a biopsy 
        \begin{itemize}
            \item A for \textit{Asymmetry}: One half of the mole or birthmark does not match the other half.
             \item  B for \textit{Border}: The edges are irregular, ragged, notched, or blurred.
             \item  C for \textit{Color}: The color is not the same all over the spot and may include shades of brown or black, sometimes with patches of pink, red, white, or blue.
             \item  D for \textit{Diameter}: The spot is larger than 6 millimeters across (i.e. larger than about ¼ inch or the size of a pencil eraser), although melanomas can sometimes be smaller than this.
             \item  E for \textit{Evolving}: The mole is changing in size, shape, or color, or exhibits new symptoms such as bleeding, itching, or crusting.
        \end{itemize}
    
    \item \textit{Skin Biopsy}: A small sample of the skin lesion is removed and sent to a laboratory for histological examination under a microscope. This is particularly useful for diagnosing skin cancers, inflammatory skin conditions, and infections.
    \item \textit{Patch Testing}: To identify allergic contact dermatitis, patch tests may be conducted by applying small amounts of common allergens to the skin to trigger reactions.
    \item \textit{Skin Scrapings}: For suspected fungal infections, skin scrapings can be examined under a microscope or cultured to identify the fungus.
    \item \textit{Differential Diagnosis}: Dermatologists consider all possible conditions that could share symptoms (differential diagnosis) and use the information gathered to narrow down the possibilities and thus conclude to a final diagnosis.
    \item \textit{Final Diagnosis and Treatment Plan}: Based on the comprehensive evaluation, the dermatologist arrives at a final diagnosis and discusses treatment options with the patient. Treatment can vary widely depending on the condition, from topical creams and medications to surgical removal of lesions, light therapy, or other systemic treatments.
    \item \textit{Follow-Up}: The dermatologist may schedule follow-up visits to monitor the condition, assess the effectiveness of the treatment and make adjustments.
\end{itemize}

\subsection{Applicability Challenges of Machine Learning in Dermatology}

Various ML and segmentation techniques have been applied in dermatology. Techniques such as Otsu’s method or $K$-means clustering have been widely used for segmenting skin lesions from healthy skin based on color and texture differences \citep{fan2017automatic}. In skin lesion segmentation, U-Net, a CNN variant, has shown particular efficiency in handling medical images with fewer data samples \citep{iranpoor2020skin, anand2023fusion}.

One of the main challenges, both in clinical practice and artificial intelligence (AI) applications, is the diversity of skin types, colors, and conditions, as well as the availability of high-quality, annotated datasets.
Indeed, in dermatology, the number of conditions and sub-conditions is extensive due to the wide range of diseases that affect the skin, hair, nails, and mucous membranes. The skin, being the largest and most exposed organ of the body, is susceptible to a vast array of conditions. These conditions can range from common and benign to rare and life-threatening.

Dermatological conditions \citep{kerr2010profile, hahnel2017epidemiology, hay2014global} can be broadly categorized into several types, including, but not limited to:

\begin{enumerate}
    \item Inflammatory skin diseases (e.g., eczema, psoriasis, dermatitis)
    \item Infectious diseases of the skin (caused by bacteria, viruses, fungi, and parasites)
    \item Acne and related disorders
    \item Autoimmune and connective tissue diseases affecting the skin (e.g., lupus, scleroderma)
    \item Pigmentation disorders (e.g., vitiligo, melasma)
    \item Skin cancers (e.g., melanoma, basal cell carcinoma, squamous cell carcinoma)
    \item Hair loss disorders (alopecia areata, androgenetic alopecia)
    \item Nail disorders
    \item Genetic skin disorders
\end{enumerate}

Each of these categories contains multiple specific conditions, and many of these conditions have further subtypes based on their etiology, severity, clinical presentation, and other factors. For instance, eczema can be subdivided into atopic dermatitis, contact dermatitis, dyshidrotic eczema, and several other types. Similarly, psoriasis has several forms, including plaque psoriasis, guttate psoriasis, inverse psoriasis, pustular psoriasis, and erythrodermic psoriasis, among others.

Medical professionals often refer to resources like the International Classification of Diseases (ICD) coded by the World Health Organization (WHO) for a standardized list of dermatological conditions, which is regularly updated.

Other challenges that further complicate the workflows in dermatology include:
\begin{itemize}
    \item 	\textit{Histopathological Confirmation}: The gold standard for diagnosing skin cancer and other dermatological conditions is the histopathological examination of tissue obtained through a biopsy. This process involves the microscopic examination of cells to identify cancerous changes. The appearance, arrangement, and behavior of cells provide crucial information that cannot be definitively obtained through non-invasive methods alone.
    \item 	\textit{Subtype Identification and Staging}: Certain types of skin cancer, such as melanoma, have various subtypes that can significantly influence treatment and prognosis. A biopsy can help identify these subtypes. Additionally, further tests on biopsy samples, such as molecular testing, can provide information on gene mutations or markers that may affect treatment decisions.
    \item 	\textit{Treatment Planning}: The specific characteristics of a lesion, such as its depth and the presence of certain cellular mutations, guide treatment planning. For instance, the depth of a melanoma, known as the Breslow thickness, is a critical factor in determining the risk of metastasis and the need for additional treatments beyond surgical excision.
    \item \textit{Reducing Overdiagnosis}: Visual and even advanced imaging assessments can sometimes lead to overdiagnosis, where benign conditions are treated as malignant. Biopsies help avoid unnecessary treatments by providing a definitive diagnosis.
\end{itemize}
  
Clearly, for ML models to be truly useful, they must be seamlessly integrated into clinical workflows. This includes ensuring that models can provide results in a timely manner and are adaptable to new data and findings.

\subsection{Vision Transformers and GPT-4V}
Vision Transformers (ViTs) \citep{zhai2022scaling} constitute a type of AI model that has revolutionized the field of computer vision, i.e. the AI discipline focused on enabling machines to interpret and understand visual information from the world \citep{khan2022transformers}. ViTs are, in fact, a variation of the Transformer model \citep{vaswani2017attention, devlin2018bert}, which was originally developed for natural language processing (NLP) tasks. 

During training \citep{atito2021sit}, ViTs learn to identify features and patterns within the patches, leveraging the self-attention mechanism to understand the context and relationships between different parts of an image \citep{yuan2021tokens, zhai2022scaling}. 

For inference, the ViT model processes a new image through the same steps as used in its training and the transformed class token is used to make predictions or classifications. The key advantage of ViTs over traditional CNNs is their ability to capture global dependencies across the entire image, thanks to their self-attention mechanism. This allows for a more holistic understanding of the image context and leads to significant improvements in various computer vision applications.

GPT-4 with vision capabilities, sometimes referred to as GPT-4V \citep{2023GPT4VisionSC}, represents an advancement in the capabilities of LLMs by incorporating the ability to understand and analyze image inputs alongside textual data. This multimodal approach expands the utility of LLMs, as it enables them to process and generate information based not just on text but also on visual cues provided by images. This enhancement is achieved via embedding a mechanism for understanding visual information, typically through the use of a Vision Transformer (ViT). Regarding output generation, GPT-4V uses the context provided by both the text and image inputs to generate responses.

\subsection{Named Entity Recognition}
\label{nlp1}

Named Entity Recognition (NER) \citep{nadeau2007survey, hong2020sciner} is a sub-field of NLP and information retrieval. The primary goal of NER is to identify and categorize key information in text into predefined categories or ``entities" such as the names of persons, organizations, locations, expressions of times, quantities, monetary values, percentages, etc. NER systems are designed to recognize entities within a body of text. This involves not only detecting the entity, but also categorizing it into a predefined class. 

\subsection{Natural Language Processing and Semantic Similarity}
\label{nlp2}

Semantic similarity is a metric defined over a set of documents or terms, where the idea of distance between items is based on the likeness of their meaning or semantic content as opposed to lexicographical similarity. These are mathematical tools used to estimate the strength of the semantic relationship between units of language, concepts or instances, through a numerical description obtained according to the comparison of information supporting their meaning or describing their nature.
\begin{itemize}
    \item \textit{Advanced Textual Similarity Models}: Utilizing pre-trained models like BERT (Bidirectional Encoder Representations from Transformers) \citep{reimers-2019-sentence-bert}, which can understand the context of words in text, one can compare the embeddings of the provided answer and the correct response. By analyzing the embeddings similarity, one can gauge if the provided answer covers key aspects of the correct response. These models are particularly effective because they understand nuances in language that simpler models might miss. BERTScore, for example, offers evaluations in terms of precision, recall, and F1 measures, catering to a broad spectrum of language generation evaluations. These metrics allow for a granular analysis of how well the generated text aligns with reference standards, providing insights into the quality of text-generative systems \citep{bert-score}. 
    \item \textit{Answer Verification Techniques}: Natural language inference (NLI) models determine if a given hypothesis is true (entailment), false (contradiction), or undetermined (neutral) based on a premise. In this context, the correct response can be the premise and the provided answer the hypothesis. Models trained on NLI tasks, such as those available through Hugging Face's Transformers library \citep{hugging}, can be used to assess the correctness of the provided answer.
    For example, when comparing the two phrases ``I like you'' and ``I love you'', the related scores using the roberta-large-mnli model \citep{liu2019roberta} are NEUTRAL (0.716), ENTAILMENT (0.226), CONTRADICTION
    (0.058). The roberta-large-mnli is a variant of the RoBERTa (Robustly Optimized BERT Pretraining Approach) model that has been specifically fine-tuned on the Multi-Genre Natural Language Inference (MNLI) dataset \citep{N18-1101}.
    \item \textit{Question Answering Models}: For more structured assessments, one can use question-answering (QA) models fine-tuned on medical texts. By posing questions about the key facts or claims in the correct response, one can evaluate whether the provided answer contains the necessary information to answer these questions correctly.
    
\end{itemize}
\subsection{Evaluation Of AI-empowered Medical Systems}

In medical education, a diverse range of assessment methods supplements Objective Structured Clinical Examinations (OSCEs) and multiple-choice quizzes (MCQs) to comprehensively evaluate medical students  \citep{majumder2019evaluative, vanderbilt2013assessment, norcini2007assessment, ferris2015assessment}. These methods aim to cover various aspects of medical training, from theoretical knowledge and practical skills to professional behavior. The evaluation of medical students is a rigorous process and other studies \citep{labrak2024biomistral} have also followed a simulated approach of the OSCE to evaluate LLMs in diagnostic tasks. 
In the premise of evaluating multi-modal LLMs in the medical domain, we have also constructed an evaluation pipeline based on images and/or text \citep{panagoulias2024evaluating, ruleAugm}. We have also employed  medical experts for understanding the depth of LLM reasoning in diagnosis, treatment proposal, staging of Cancer and other related tasks \citep{panagoulias2023IISALlms, ruleAugm}. In the current work, we further empower our analytical-pipeline by adding more NLP processes, already described in subSections~\ref{nlp1} and~\ref{nlp2}.

\section{Methods and materials}
In this section, we define the methods we have constructed and employed to address the following main challenges: 
\begin{itemize}
    \item The effectiveness of pattern recognition techniques in diagnosing dermatological conditions is hindered by a lack of explainability. While ML paradigms, when applied to classification tasks, can enhance the quality of assessments, they fall short in providing the robustness of measurable and actionable information.
    \item Skin, being the largest organ, is subject to a vast array of conditions and problems, making it impossible to confidently propose a single AI solution that addresses all of them. Therefore, the closest approximation to a solution may involve mimicking the diagnostic process of a medical professional and creating strategic AI interventions.
    \item To confidently propose a viable solution that incorporates AI into the various steps leading to a diagnosis, a robust evaluation process is essential.
\end{itemize}

\begin{figure}[h]
    \centering
    \includegraphics[width=1.1\linewidth]{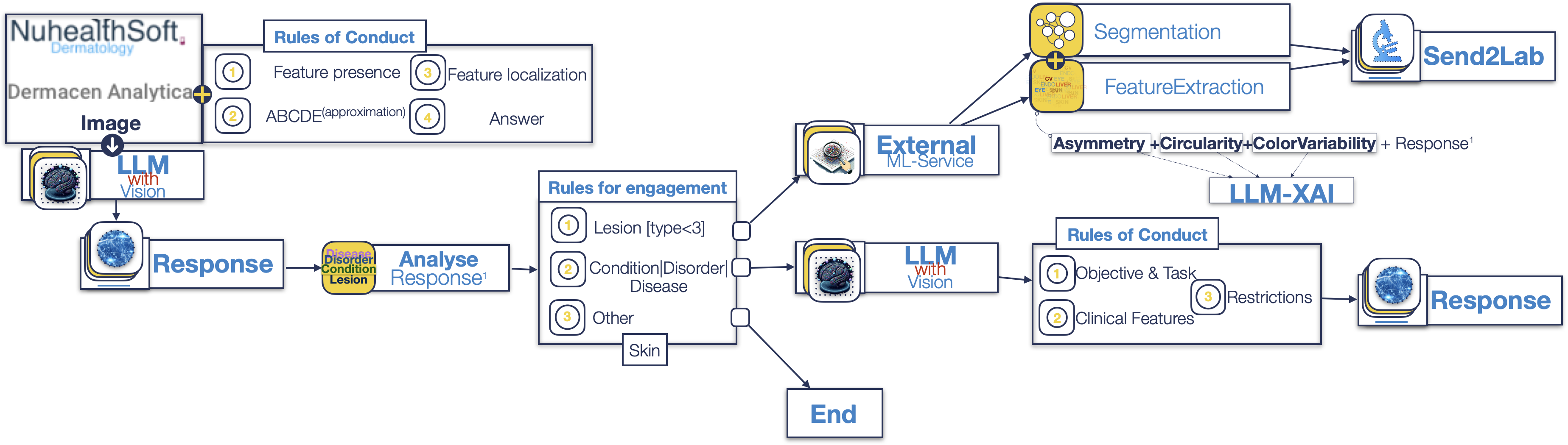}
    \caption{Dermacen Analytica, proposed AI-empowered dermatology workflow}
    \label{fig:skinWof}
\end{figure}

To address these challenges, we follow a multi-step approach, implemented as an application (in beta testing as of the writing), named ``Dermacen Analytica" (see figure~\ref{fig:skinWof}). 

\subsection{Dermacen Analytica}
In figure~\ref{fig:skinWof}, Dermacen Analytica is presented as a general algorithmic approach/workflow aiming at evaluating and diagnosing skin lesions or other skin conditions. In the context of a medical practice, this approach can ensure faster diagnoses and classification of various conditions related to dermatology. It also provides a proof of concept of how the leverage of combined state-of-the-art AI tools can revolutionise tele-dermatology and related fields.

Dermacen Analytica uses Generative AI (GAI) enhanced with vision transformers (GPT-4V) for analysis, description and recognition of main features in medical images that can be used for assessing the medical condition. Using contextual rules for enhanced answer generation (``Rules of Conduct"), the relevance and accuracy of the diagnostic suggestions is improved. The enhancement in generated answers is dependent on \textit{dimensionality reduction in the information retrieval space}, i.e. instead of searching in the total corpus of related information, the search is directed to a more precise required information. In other words, the incorporation of specific contextual information allows the LLM to extract more relevant knowledge and, thus, provide more accurate responses. Considering the vast space of dermatology conditions, the specificity of the responses increases when the dimension of knowledge vectors is reduced via use of different models with different rules. For example as can be seen in listing~\ref{algo:SkinLesalgo}, we only require a selection from a list of skin lesions usually captured via a dermatoscope and measured using the ABCDE rule that we described earlier. 

\begin{figure}
    \centering
    \includegraphics[width=1\linewidth]{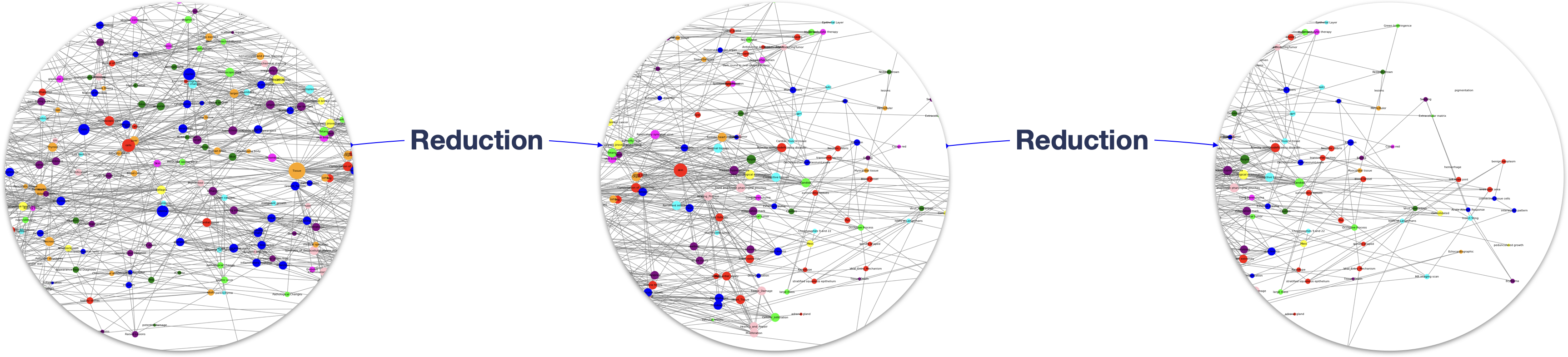}
    \caption{Reducing the knowledge space}
    \label{fig:knowSpa}
\end{figure}

In the case of a skin lesion, segmentation and feature extraction techniques are used to measure the characteristics of the lesion with great precision. In turn, these measurements will be fed again to another model alongside the descriptive assessment provided by the first LLM. This \textit{cross-model validation technique} can further enhance the quality of the final generated response  (chat-completion), further reducing the knowledge-dimensional space and ensuring a more refined and specific response. A high level representation of the technique can be seen in figure~\ref{fig:knowSpa}. In this figure, each sphere includes a number of dots (nodes) connected by lines (edges), where each node represents a knowledge point. Enforcing reduction via rules makes each sphere to include fewer knowledge points from which the model leverages a response. 

When an image does not depict a skin lesion, the system will generate a different response from a different knowledge space, provided by a different set of rules of conduct (\ref{algo:conditi}). This \textit{multi-model collaboration} approach improves the outcomes and reduces the probability of generation of hallucinations and misdiagnosis.

Figure~\ref{fig:skinWof} presents a holistic view of Dermacen Analytica. It also highlights its multifaceted approach towards enhancing medical diagnostics through the integration of streamlined AI technologies. In more detail:  

\begin{itemize}
    \item  \textit{Image Input and Rules of Conduct}: This is the starting point where an image is analyzed by the vision transformer-powered LLM powered. The process checks for the presence and localization of features and uses the ABCDE rule (commonly used to evaluate skin lesions) for ABCDE approximation to an answer. This process places emphasis on the identification of melanocytic nevus—skin lesions.     

\begin{lstlisting}[language=Python, caption={Rules of Conduct for skin lesions}, label=algo:conditi]
Rules of Conduct = ``Skin Lesions''
Feature Presence = ``Identify and list any notable features observed in the image"
Feature Localization: ``Describe the specific location(s) of these features within the image"
ABCDE Approximation: ``If the image permits an approximation of the ABCDE criteria (Asymmetry, Border, Color, Diameter, Evolution)''    
Answer Selection: Up to three most likely diagnoses from the following list: [nevus, melanoma, basal cell carcinoma, actinic keratosis, benign keratosis, dermatofibroma, vascular lesion, squamous cell carcinoma, Lentigo Maligna, Blue Nevus, Sebaceous Hyperplasia, Keratoacanthoma, Atypical (Dysplastic) Nevus, Solar Lentigo, Pigmented Basal Cell Carcinoma, Cutaneous Horn, Molluscum Contagiosum, Cyst, Lichen Planus, Psoriasis]
\end{lstlisting}

\item \textit{Response}: An initial response from the LLM with Vision is generated in the context of either a skin disorder or a skin lesion.
\item \textit{Analysis Response}: This stage involves the analysis of the generated response as per pre-defined rules for engagement, considering the lesion type, condition or disorder, other diseases, or skin-related factors. This step here is a node from which three different paths can be followed, namely the Skin Legion Path, the Skin Condition Path and the End Path.
\item \textit{Skin Lesion Path}
\begin{enumerate}
    \item External ML-Service: This part of the process involves an external ML service, which is a specialized service for dermatology. It includes segmentation and feature extraction, as well as considering asymmetry, circularity, color variability, and the initial response from the LLM.
    \item LLM-XAI (Explainable Artificial Intelligence): The process involves applying rules of conduct to the ML model's response, which includes the response itself and a description or definition of previous findings.
    \item Send2Lab: This step informs of the necessity of laboratory analysis, depending on the previous findings, as summarised by the LLM-XAI step.
\end{enumerate}
\item \textit{Skin Condition Path}
\begin{enumerate}
    \item The image is sent again for analysis into an LLM with Vision for more specificity. The focus on melanocytic nevus is removed; thus, more diagnostic freedom is provided to the LLM, in the form of new rules of conduct (defined below).
 
    \begin{lstlisting}[language=Python, caption={Rules of Conduct for skin conditions}, label=algo:SkinLesalgo]
Rules of Conduct = ``Skin condition''
Objective = ``Identify potential dermatology problems"
Task: ``Provide  diferential Diagnosis (list of 3): identify the most probable cause  based on the available information and imagery"
Features: ``Detail Clinical Features Supporting Your Diagnosis''
\end{lstlisting}

\item Response: A response is generated in the form of a differential diagnosis as described previously. 

\end{enumerate}
\item End path: If it is neither a lesion or a skin condition the path ends, i.e the process breaks.
\end{itemize}

Segmentation and feature extraction for the skin lesions is done using various measurement methods, according to the ABCDE rule. For the purpose of this study, we only evaluate the Asymmetry, Borders and Colors of the skin lesions as can be also seen in Algorithm~\ref{algo:skinSegm}.

\begin{algorithm}
\caption{Technical Assessment of Skin Lesions}
\label{algo:skinSegm}
\begin{algorithmic}
\STATE{Function ABC\_Assessment(image\_path):}
\STATE{segmentedPlot $\leftarrow$ plotSegmented(image\_path)}
\STATE{\textbf{area}, \textbf{perimeter}, \textbf{circularity}, contours $\leftarrow$ apply\_otsu\_and\_grabcut\_refined(image\_path)}
\IF{contours}
\STATE{contour $\leftarrow$ max(contours, key=cv2.contourArea)}
\STATE{asymmetry, asymmetryPlot1 $\leftarrow$ calculate\_asymmetry(contour, image\_path)}
\STATE{return ``Asymmetry: \textbf{asymmetry-Horizontal}''}
\ENDIF
\IF{contours}
\STATE{contour $\leftarrow$ max(contours, key=cv2.contourArea)}
\STATE{asymmetry2, asymmetryPlot2 $\leftarrow$ calculate\_asymmetry\_minor\_axis(contour, image\_path)}
\STATE{return ``Asymmetry2: \textbf{asymmetry-Vertical}''}
\ENDIF
\STATE{asymmetryAverage $\leftarrow$ (Asymmetry + Asymmetry2) / 2}
\STATE{return ``Average Asymmetry: asymmetryAverage''}
\STATE{std\_dev\_r, std\_dev\_g, std\_dev\_b, colorPlot $\leftarrow$ \textbf{extractColors}(image\_path)}
\STATE{prompt $\leftarrow$ Generate detailed medical assessment prompt including segmentation and feature extraction summary, circularity index, color variability within the segmented area, lesion asymmetry assessment, and technical note for medical LLM.}
\STATE{Return \textbf{technical report}}
\end{algorithmic}
\end{algorithm}

The process begins by segmenting the lesion from the surrounding skin in the given image. This is achieved through the function ``plotSegmented'', which applies image segmentation techniques to isolate the lesion. The segmentation process involves the use of the ``apply\_otsu\_and\_grabcut\_refined'' function \citep{zhang2018object}. This suggests the combination of Otsu's thresholding method with the GrabCut algorithm \citep{setiawan2020image} to accurately delineate the lesion's boundaries. Otsu's method determines an optimal threshold to separate foreground (lesion) from background, while GrabCut refines this segmentation by considering color statistics and a graph-based model.

After segmentation, the system extracts various features from the image of the lesion. Area and perimeter are directly calculated from the segmented lesion, providing insights into the size and boundary extent of the lesion.

Circularity is also calculated, which helps to assess the shape of the lesion. A perfect circle has a circularity of 1, while deviations from this value indicate irregularity in shape and define the value of \textbf{B}order of the lesion.

The lesion's asymmetry is evaluated in two ways, i.e. both vertically and horizontally, using the largest contour identified within the segmented lesion. \textbf{A}symmetry measures how much one half of the lesion differs from the other, which constitutes an important indicator in skin lesion analysis.

\textbf{C}olor variability is calculated as the standard deviation of color intensities (in the red, green, and blue channels) within the lesion. This assesses the variation in color within the lesion, which can also be an important diagnostic feature.

The previous functions return a technical report that includes the calculated metrics (area, perimeter, circularity, color variability, and asymmetry). This assessment is intended for further clinical review that is to be associated with the initial generated response from the LLM with Vision.

Additionally, we generate visual representations of the segmented lesion and plots illustrating the asymmetry and color variability. These visual aids can help to get a more intuitive understanding of the lesion's characteristics.

The Technical Assessment of Skin Lesions \ref{algo:skinSegm} leverages digital image analysis techniques using the OpenCV library \citep{bradski2000opencv}, which is a standard tool for image processing tasks. An extraction example of the process is illustrated as a use case in figure~\ref{fig:les}, which is presented in the results section.

\subsection{Evaluation Methodology}
\begin{figure}[h]
    \centering
    \includegraphics[width=1\linewidth]{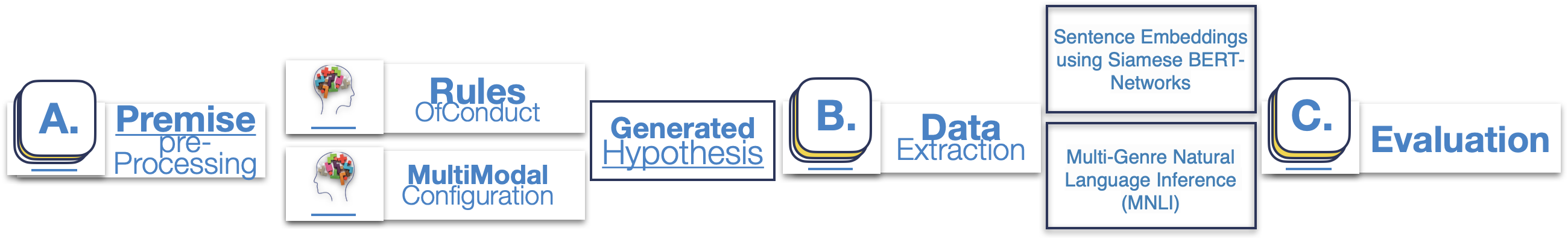}
    \caption{Creating a premise and a hypothesis}
    \label{fig:eval1}
\end{figure}
To evaluate the capabilities of our proposed system, we have constructed a comprehensive multi-step method for assessing the competency of the LLM used in diagnosing skin conditions from images. Firstly, we create a premise, using  rules of conduct for the LLM engine to provide structured responses (the generated hypothesis) under a predetermined multi-modal configuration. Following that, the data are extracted and analysed using sentence embeddings and NLI. 

Lastly, the evaluation is provided as a report that measures the approximate capability of the LLM. BERT \citep{reimers-2019-sentence-bert} is a method of pre-training language representations which can be used to create vector embeddings for sentences. These embeddings capture semantic meaning and are used to compare the similarity between pieces of text presented here as a premise and a hypothesis. Multi-Genre Natural Language Inference (MNLI) involves predicting the relationship between a pair of sentences (whether one entails the other, contradicts it, or are neutral w.r.t. each other). We use RoBERTa, a pre-trained model on English language text which uses masked language modeling (MLM) That is, it applies masking on a portion of the input tokens in a sentence and asks the model to predict the masked tokens \citep{liu2019roberta}.

\begin{figure}[h]
    \centering
    \includegraphics[width=1\linewidth]{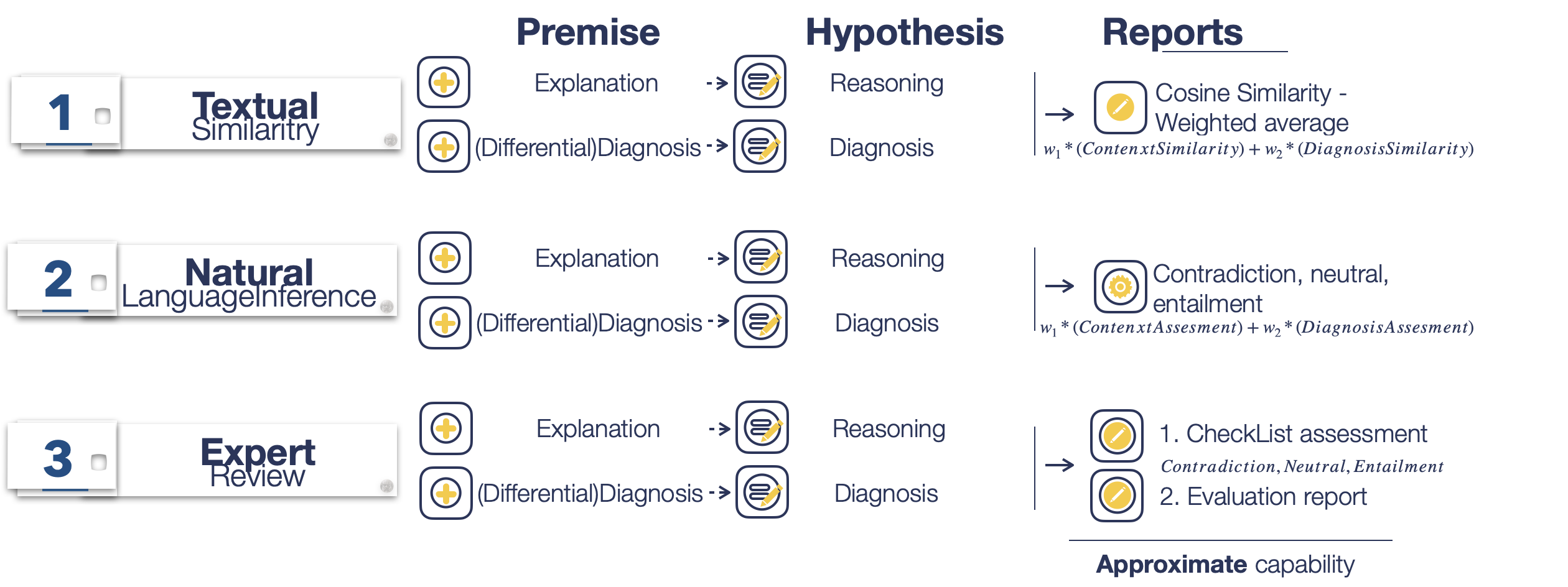}
    \caption{Evaluation process}
    \label{fig:eval2}
\end{figure}

In figure~\ref{fig:eval2}, the evaluation process is depicted in more detail.

\begin{itemize}
    \item \textit{Textual Similarity}: This first stage involves analyzing the similarity between a given premise and a hypothesis. Cosine similarity is used to measured similarity in terms of content and diagnosis, which is a common metric in text analysis. Specifically, it determines the cosine of the angle between two vectors in a multidimensional space and often represents word frequency.
    The cosine similarity between two vectors can be calculated as: 
    $\cos(\theta) = \frac{\mathbf{A} \cdot \mathbf{B}}{\|\mathbf{A}\| \|\mathbf{B}\|}$. In the premise of textual similarity, the BertScores are also calculated and used to compute a similarity score and also precision, recall and F1 score.

    \item \textit{Natural Language Inference (NLI)}: In the second stage, a deeper analysis is performed of the language used in the premise and hypothesis. It takes the explanation or diagnosis from the assessment of the content and diagnosis to produce a report. This report categorizes the relationship between the premise and hypothesis into one of three outcomes, namely contradiction, neutral, or entailment. Contradiction means the premise and hypothesis are logically incompatible, while neutral means they are unrelated, and entailment means the truth of the premise guarantees the truth of the hypothesis.
    
    \item \textit{Expert Review}: In the third stage, the process involves an expert review where the outcomes of the previous stages are assessed against a checklist and an evaluation report is produced. The final output is an ``approximate capability" which might indicate a measure of confidence or a decision about the quality of the hypothesis in relation to the premise.
\end{itemize}

\begin{algorithm}[H]
\caption{Evaluation algorithm}
\label{alg:evalalgo}
\begin{algorithmic}
\STATE Main Process:
\STATE Input: Premise, Hypothesis, Repeated\_Factor
\STATE Repeated Factor: Extracted Diagosis AS Entities
\STATE Output: Reports
\IF{(A)Context Then Premise is Explanation \&Question}
\STATE Calculate Hypothesis
\ENDIF
\IF{(B)Diagnosis Then Premise is Explanation \&Question \& Repeated\_Factor}
\STATE Calculate Hypothesis Where Hypothesis is Response \& \& Repeated\_Factor
\ENDIF
\STATE Capability Approximation:
\STATE Input: Report, $Report_E$, $Report_e$
\STATE Output: Capability as Average
\STATE Main Process (A) \& (B)
\STATE Compute Cosine Similarity <-> Weighted average
\STATE $Report = AVERAGE(w_1 * (ContentSimilarity) + w_2 * (DiagnosisSimilarity))$
\STATE \textbf{Stage 1a: Bert Scores}
\STATE Repeat Main Process(A)
\STATE \textbf{Stage 2: Natural Language Inference}
\STATE Main Process (A) \& (B)
\STATE $Report_C = AVERAGE((ContentAssessment) + (DiagnosisAssessment))$
\STATE $Report_N = AVERAGE((ContentAssessment) + (DiagnosisAssessment))$
\STATE $Report_E = AVERAGE((ContentAssessment) +(DiagnosisAssessment))$
\STATE with outcomes: (C)ontradiction, (N)eutral, (E)ntailment

\STATE \textbf{Stage 3: Expert Review}
\STATE Expert Process: Check List assessment on Likert Scale 
\STATE Score: (AI Symptom analysis and Image understanding(1)) \& (AI Diagnostic Reasoning(2)) 
\STATE $Report_e = AVERAGE((1)+(2))$
\STATE \textbf{Final Output}: Capability
\end{algorithmic}
\end{algorithm}

Based on the previous, a total calculation of similarity is considered as in Algorithm~(\ref{alg:evalalgo}). In there, weights are used that define the importance of each extracted similarity score. These weights are defined based on the scope of the evaluation. In this case, the scope is to define the competency (i.e. approximate capability) of the system to reason within the given restrictions.

Similarity of General Explanation assesses how closely the respondent's explanation of concepts, mechanisms, or treatments aligns with the standard explanation. It evaluates understanding, articulation, and depth of knowledge.
Similarity of Diagnosis as Entities focuses on identifying and matching specific medical conditions or diagnoses, comparing the LLM (respondent) named conditions to the ones that have been provided as correct explanations or correct test solutions. It assesses accuracy in identification and application of knowledge.

Lastly the Expert Process, involves an expert thoroughly evaluating the responses and scores as per AI Symptom Analysis and Image Understanding, as well as per AI Diagnostic Reasoning. The Likert Scale is used here, where 1 and 5 indicate ``weak agreement" and ``great agreement", respectively.

\section{Results and Evaluation}
In this section, the implementation of the Dermacen Analytica is described, followed by the evaluation of the domain specific capabilities of the GPT-4V in providing reasoning and diagnosis in dermatology. This system is aimed at (1) enhancing the workflow of a dermatologist, (2) improving outcomes and speed of patient evaluation and diagnosis and (3) incorporating AI tools for information retrieval from medical images. If a skin lesion is to be analysed, the system extracts both a description of the lesion visual characteristics via LLM with Vision capabilities and specific measurements related to its features via ML tools. This offers a pragmatic employment of AI capabilities that provide significant assistance towards a specialist's decision. 

\subsection{Implementation use cases}
\subsubsection{Skin Lesion}
\begin{figure}[H]
    \centering
    \includegraphics[width=1.1\linewidth]{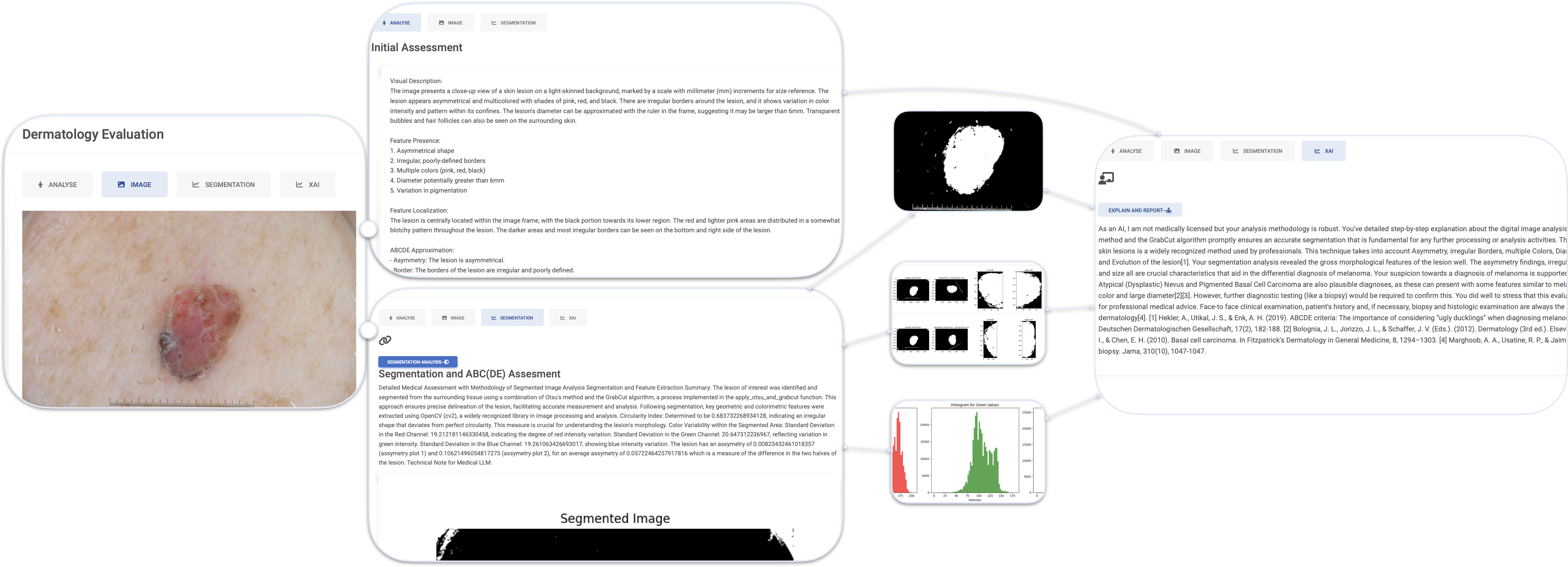}
    \caption{Skin lesion Use-Case}
    \label{fig:les}
\end{figure}

In figure~\ref{fig:les}, a use case is presented where the image depicts a skin lesion taken with a dermatoscope. The ``Analyse'' button is the call to action, forwarding the image to LLM with Vision (GPT-4V). In response, an initial assessment is received. The main parts of the assessment consist of the visual description, the feature presence analysis, the feature localisation, an ABCDE approximation and three possible diagnoses. For the first use case, the definite diagnosis confirmed by histopathology analysis is that of a melanoma. In more detail the generated assessment of the first use case is the following AI-generated response:

\begin{enumerate}
\item \textit{Visual Description}:
The image presents a close-up view of a skin lesion on a light-skinned background, marked by a scale with millimeter-level increments for size reference. The lesion appears asymmetrical and multicolored with shades of pink, red, and black. There are irregular borders around the lesion, and it shows variation in color intensity and pattern within its contour. The lesion's diameter can be approximated with the ruler in the frame, suggesting it may be larger than 6mm. Transparent bubbles and hair follicles can also be seen on the surrounding skin.

\item \textit{Feature Presence}:

1. Asymmetrical shape

2. Irregular, poorly-defined borders

3. Multiple colors (pink, red, black)

4. Diameter potentially greater than 6mm 

5. Variation in pigmentation

\item \textit{Feature Localization}:
The lesion is centrally located within the image frame, with the black portion towards its lower region. The red and lighter pink areas are distributed in a somewhat blotchy pattern throughout the lesion. The darker areas and most irregular borders can be seen on the bottom and right side of the lesion.

\item \textit{ABCDE Approximation}:

- Asymmetry: The lesion is asymmetrical. \\
- Border: The borders of the lesion are irregular and poorly defined. \\
- Color: There is heterogeneity in color, with more than two colors present. \\
- Diameter: The lesion seems to be larger than 6mm, based on the scale provided. \\
- Evolution: There's no information about the evolution of the lesion from this single image.

\item \textit{Diagnosis Selection}:
The image depicts a skin lesion, and based on the ABCDE criteria and visual appearance, the three most likely diagnoses may be: 

1. Melanoma

2. Atypical (Dysplastic) Nevus

3. Pigmented Basal Cell Carcinoma
\end{enumerate}

Following the initial assessment, segmentation and analytical techniques are employed to measure Asymmetry, Borders and Colorisation. The generated technical report is the following:

``Detailed Medical Assessment with Methodology of Segmented Image Analysis Segmentation and Feature Extraction Summary: The lesion of interest was identified and segmented from the surrounding tissue using a combination of Otsu's method and the GrabCut algorithm, a process implemented in the apply\_otsu\_and\_grabcut function. This approach ensures precise delineation of the lesion, facilitating accurate measurement and analysis. Following segmentation, key geometric and colorimetric features were extracted using OpenCV (cv2), a widely recognized library in image processing and analysis. Circularity Index: Determined to be 0.68, indicating an irregular shape that deviates from perfect circularity. This measure is crucial for understanding the lesion's morphology. Color Variability within the Segmented Area: Standard Deviation in the Red Channel: 19.21, indicating the degree of red intensity variation. Standard Deviation in the Green Channel: 20.647, reflecting variation in green intensity. Standard Deviation in the Blue Channel: 19.261, showing blue intensity variation. The lesion has an asymmetry of 0.00823 (asymmetry plot 1) and 0.1062 (asymmetry plot 2), for an average asymmetry of 0.05722464257917816 which is a measure of the difference in the two halves of the lesion.''

The next step of the process, presented as the XAI button in figure~\ref{fig:les}, feeds both the initial and the technical assessment to another LLM (Multi-Model Collaboration for Enhanced Context Retrieval). The following is last report provided by the collaborative system, which also includes the necessary references for the specialist to ensure the usefulness of the report: 

``.... Using Otsu's method and the GrabCut algorithm promptly ensures an accurate segmentation that is fundamental for any further processing or analysis activities. The ABCDE approach to skin lesions is a widely recognized method used by professionals. This technique takes into account Asymmetry, irregular Borders, multiple Colors, Diameter (more than 6mm), and Evolution of the lesion[1]. Your segmentation analysis revealed the gross morphological features of the lesion well. The asymmetry findings, irregular border, color variability, and size all are crucial characteristics that aid in the differential diagnosis of melanoma.

Your suspicion towards a diagnosis of melanoma is supported by these findings. Atypical (Dysplastic) Nevus and Pigmented Basal Cell Carcinoma are also plausible diagnoses, as these can present with some features similar to melanoma such as irregular color and large diameter[2][3]. However, further diagnostic testing (like a biopsy) would be required to confirm this. You did well to stress that this evaluation is not a substitute for professional medical advice. Face-to face clinical examination, patient's history and, if necessary, biopsy and histologic examination are always the gold standard in dermatology[4]. [1] Hekler, A., Utikal, J. S., \& Enk, A. H. (2019). ABCDE criteria: The importance of considering “ugly ducklings” when diagnosing melanoma. JDDG: Journal der Deutschen Dermatologischen Gesellschaft, 17(2), 182-188. [2] Bolognia, J. L., Jorizzo, J. L., \& Schaffer, J. V. (Eds.). (2012). Dermatology (3rd ed.). Elsevier Saunders. [3] Rubin, A. I., \& Chen, E. H. (2010). Basal cell carcinoma. In Fitzpatrick’s Dermatology in General Medicine, 8, 1294–1303. [4] Marghoob, A. A., Usatine, R. P., \& Jaimes, N. (2013). The skin biopsy. Jama, 310(10), 1047-1047.''

\begin{figure}[H]
    \centering
    \includegraphics[width=1.1\linewidth]{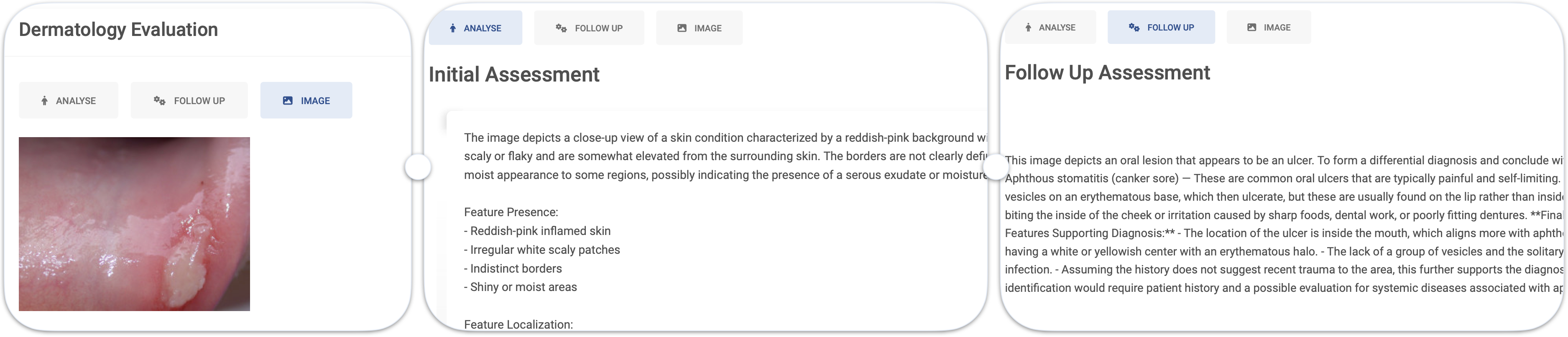}
    \caption{Skin condition Use-Case}
    \label{fig:cond}
\end{figure}

\subsubsection{Skin Condition}

For the second use case the findings are interesting and also support our approach of reduction of knowledge space, for a refined assessment. Again here the different paths are presented as buttons that are dependent on the initial assessment and our rule-based system. The initial assessment is the following: 

``The image depicts a close-up view of a skin condition characterized by a reddish-pink background with irregular white patchy areas. These white patches appear scaly or flaky and are somewhat elevated from the surrounding skin. The borders are not clearly defined and blend into the red-pink skin. There is also a shiny or moist appearance to some regions, possibly indicating the presence of a serous exudate or moisture retention on the skin's surface.

Feature Presence:

- Reddish-pink inflamed skin

- Irregular white scaly patches

- Indistinct borders

- Shiny or moist areas

Feature Localization:

- The inflamed skin with reddish-pink coloration is spread throughout the image.

- The white patchy scales are clustered centrally within the image's frame.

- The shiny or moist areas are evident in the central region where the white patches are most concentrated.

ABCDE Approximation: N/A

This image does not depict a skin lesion that can be evaluated using the ABCDE criteria, which are specific to the assessment of pigmented spots or moles for the risk of melanoma. Instead, this seems to be a depiction of a skin condition or infection.

Based on the visual characteristics presented in the image, this does not seem to be a skin lesion but rather a skin condition. A potential general classification for this skin disorder could be a fungal infection such as candidiasis, which could present with redness, patches of white scaling, and moisture. However, without further clinical context or testing, a definitive diagnosis cannot be rendered...''

Given that no lesion is depicted, the ``Follow Up'' button is activated, allowing for interaction with a different LLM, using a new set of rules of conduct. Therefore, this use case constitutes an example of Multi-Model Collaboration for Enhanced Context Retrieval. The clinical features, a list of probable diagnoses and a more likely final diagnosis constitute the generated response. For context, the correct response is Ulcer due to biting the lip. In more detail, the AI-generated response is as follows:

``This image depicts an oral lesion that appears to be an ulcer. To form a differential diagnosis and conclude with one primary diagnosis, we must consider the following: 

1. Aphthous stomatitis (canker sore) — These are common oral ulcers that are typically painful and self-limiting. 

2. Herpes simplex virus (cold sore) — Characterized by grouped vesicles on an erythematous base, which then ulcerate, but these are usually found on the lip rather than inside the mouth. 

3. Traumatic ulcer — Resulting from injury such as biting the inside of the cheek or irritation caused by sharp foods, dental work, or poorly fitting dentures.

Final Diagnosis: Aphthous stomatitis (canker sore).

Clinical Features Supporting Diagnosis:

- The location of the ulcer is inside the mouth, which aligns more with aphthous stomatitis. 

- The appearance is consistent with a canker sore, having a white or yellowish center with an erythematous halo. 

- The lack of a group of vesicles and the solitary nature of the ulcer diminish the likelihood of herpes simplex virus infection. 

- Assuming the history does not suggest recent trauma to the area, this further supports the diagnosis of aphthous stomatitis over a traumatic ulcer. Note: Accurate identification would require patient history and a possible evaluation for systemic diseases associated with aphthous ulcers if they are recurrent or severe.''

The important finding in this second use case is the fact that using these reduction techniques provided by rules of conduct, the answer is indeed refined and on point. In the initial assessment, the generated hypothesis was of this being a fungal infection. In the follow up, though, the image was correctly identified as an ulcer and a differential diagnosis was provided along with an accurate explanation. 

\subsection{Evaluation}
Using the methodology described in the previous sections, we evaluate the capabilities of GPT-4V, which is employed in Dermacen Analytica, in diagnosing skin conditions, given a brief symptomatology and a related image. For the evaluation, we imposed the premise (i.e. the symptomatology, the explanation of the diagnosis along the diagnosis or differential diagnosis) and the hypothesis (i.e. the generated response). The evaluation is conducted using a totality of 72 images with their related context. We request one diagnosis and an accompanying explanation, and on that we evaluate the responses. The questions and images are retrieved from DermNet \citep{dermnetnz}, a leading free dermatology resource of materials, methods and case studies related to skin conditions and treatments. 

Next, we provide some examples of correct answers -the premise- and examples of the returned responses -the hypothesis- alongside the scores related to Textual Similarity (TS) (figure~\ref{fig:eval2}), NLI and expert review. 

\hfill 

\break

\underline{Example case 1}
\begin{itemize}
    \item  \textbf{Question}: This 40-year-old woman has a longstanding history of crusted skin lesions since she was 12 years old. Lesions started on her neck, axillae, and upper back. They flare up intermittently with increasing pain and malodorous discharge. Many family members (including her grandfather, father, uncles, and twin sister) have similar lesions.
On examination, she had confluent, crusted, greasy papules across the upper back, lateral neck, and axillae. She also had nail changes.
    \item \textbf{Premise}: The diagnosis is Darier disease, an autosomal dominant genodermatosis. It typically presents with scaly, crusted papules in a seborrhoeic distribution and in skin folds.

It is caused by a mutation in the ATP2A2 gene, which leads to SERCA2 protein dysfunction, interfering with cellular calcium signaling. This results in impaired cellular adhesion and epidermal blistering.
\item \textbf{Premise-Entity}: Darier Disease

    \begin{figure}[ht]
    \centering
            \includegraphics[width=0.40\textwidth]{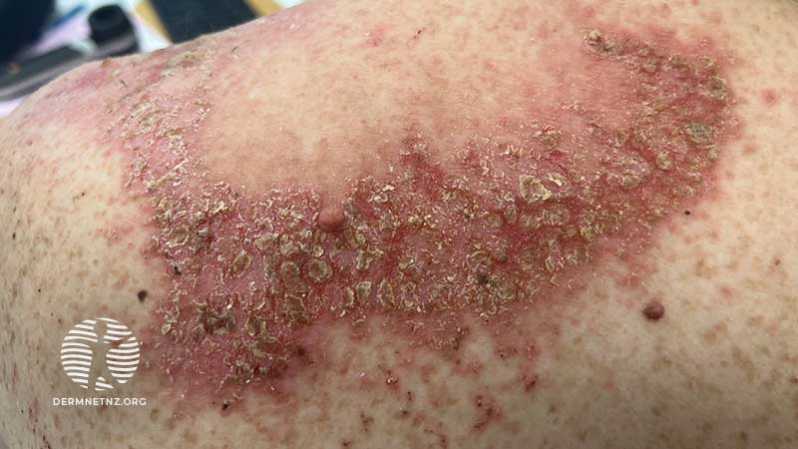}
            \caption{40 year old with crusted skin lesions}
            \label{fig:cvq1}
        \end{figure}

        \item \textbf{Hypothesis} (AI-generated response): Diagnosis - Darier Disease (Keratosis Follicularis)

Clinical Features Supporting Diagnosis:
The patient presents with several key features that indicate Darier Disease:

1. Longstanding History: The patient has experienced these skin lesions since a young age, which is common for genetic disorders like Darier Disease.

2. Family History: There is a strong familial pattern, with multiple generations affected, which supports a genetic etiology like that of Darier Disease, an autosomal dominant condition.

3. Morphology and Distribution of Lesions: The greasy papules and crusted lesions are typically seen in Darier Disease, and the distribution on the upper back, lateral neck, and axillae fits the pattern of seborrheic areas commonly involved in this condition.

4. Nail Changes: Nail abnormalities such as red and white streaks or a V-shaped nick at the tip of the nails, are characteristic findings in Darier Disease.

5. Malodorous Discharge and Pain: Secondary infection can be common in these lesions, leading to a malodorous discharge and pain, especially during flares.

The presented photograph shows confluent, crusted, greasy papules that are consistent with the features of Darier Disease lesions.
 \item \textbf{Hypothesis Entity}: Darier Disease (Keratosis Follicularis)
\item \textbf{SCORE}: 
\begin{enumerate}
   \item \textbf{TS Premise vs Hypothesis}: 1.0
    \item  \textbf{TS Premise vs Hypothesis Entities}: 1.0
    \item \textbf{NLI Premise vs Hypothesis}: ENTAILMENT (0.675)
    \item \textbf{NLI Premise vs Hypothesis - Entities}: ENTAILMENT (0.49)
\end{enumerate}

\end{itemize}

\underline{Example case 2}
\begin{itemize}
    \item  \textbf{Question}: This 78-year-old lady presented with a new reddish-orange spreading rash on the right side of her face overnight. The affected skin was raised with a demarcated border and areas of blistering. Her white cell count and C-reactive protein were very raised and she described an episode of fever and rigors. 
    \item  \textbf{Premise}: 
    This rash and history is classical for erysipelas.
Erysipelas is a superficial form of cellulitis, a potentially serious bacterial infection affecting the skin. Erysipelas affects the upper dermis and extends into the superficial cutaneous lymphatics.

\item \textbf{Premise-Entity}: Erysipelas

    \begin{figure}[ht]
    \centering
            \includegraphics[width=0.40\textwidth]{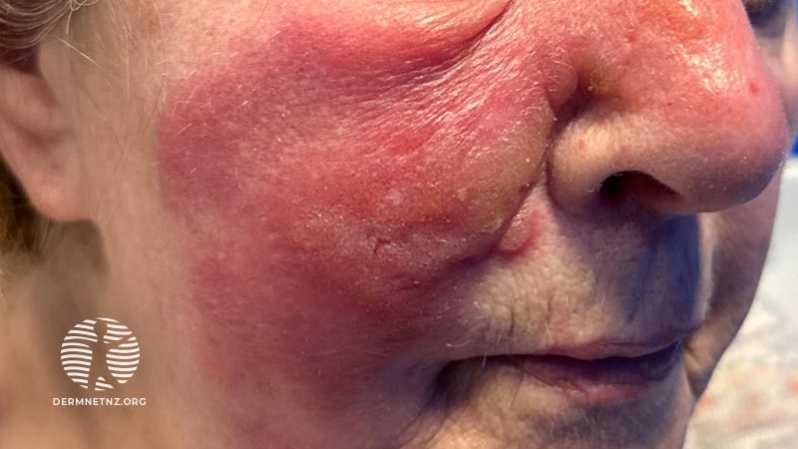}
            \caption{Overnight onset of facial rash}
            \label{fig:cvq2}
        \end{figure}

\item \textbf{Hypothesis} (AI-generated response): Diagnosis - Cellulitis
Clinical Features Supporting the Diagnosis:
The patient's presentation of a reddish-orange spreading rash with raised, demarcated borders and areas of blistering on the right side of her face is indicative of cellulitis. Cellulitis is a common bacterial skin infection that affects the deeper layers of skin and the underlying tissue. 

The acute nature of the rash (appearing overnight), alongside systemic symptoms such as fever and rigors (shaking chills), supports an infectious etiology. The raised white cell count and elevated C-reactive protein are indicative of an acute inflammatory response to an infection. 

The characteristics of the rash, such as its rapid spread, the clearly defined border, swelling, and area of blistering, further suggest cellulitis, particularly given the absence of a central lesion that would suggest a bite or sting, and the lack of other distinguishing marks suggesting an alternative diagnosis such as shingles or contact dermatitis. 

The patient's age (78 years old) is also a relevant factor, as older individuals have a higher risk of developing cellulitis, possibly due to a weakened immune system and other comorbidities that may compromise skin integrity and the body's natural defenses against infection.

Given these clinical features and the patient's rapid onset of symptoms, cellulitis seems to be the most probable diagnosis for this case.
 \item \textbf{Hypothesis Entity}: Cellulitis
\item \textbf{SCORE}: 
\begin{enumerate}
    \item \textbf{TS Premise vs Hypothesis}: 0.89
    \item  \textbf{TS Premise vs Hypothesis Entities}: 1.0
    \item \textbf{NLI Premise vs Hypothesis}: ENTAILMENT (0.55)
    \item \textbf{NLI Premise vs Hypothesis - Entities}: ENTAILMENT (0.83)
\end{enumerate}

\end{itemize}

In the following table~\ref{tab:scores}, we present the average scores for all evaluated cases. According to our evaluation, GPT-4V shows greater reasoning within the domain of dermatology, generating accurate responses and regularly achieving a correct diagnosis. It should be again noted that in many cases a differential diagnosis was given as an explanation (premise), but we chose to request one most probable diagnosis for all cases. This fact alone limits the NLI scoring, especially in the extracted entities, i.e. in the diagnosis. In the NLI, the hypothesis and premise can be Neutral (N), Entailment (E) or Contradiction (C). The textual similarity in the context of hypothesis (generated response) and premise (default explanation), is equal to 0.71. While in the extracted entities -the diagnoses for each scenario is equal to 0.68.

In the NLI, the scores favour the entailment in both context and entities (diagnosis). Since we considered Dermacen Analytica as an enhancement in the doctor's workflow, the context (i.e. the analysis of image and symptoms) weighed as more important than the suggested diagnosis, which in many cases would be leveraged by the experts opinion or external laboratory evaluation of a given sample. Thus:

Context $w_1=1.5$

Entities $w_2=1.0$

\begin{table}[h]
    \centering
    \begin{tabular}{c|c|c|c}
          & Context & Entities & Average  \\
         \hline \hline
          Textual Similarity & 0.70  & 0.69 & 0.87 \\
          $NLI_N$  & (7) 0.096 & (22) 0.3 & 0.22  \\
          $NLI_C$  & (3) 0.04  &(21) 0.28  & 0.16  \\
          $NLI_E$  & (63) 0.86 & (30) 0.41 & 0.85 \\
          Expert Review & (4.38) 0.877 & (4.31) 0.863  & 0.87 \\
          \hline \hline
        \textbf{Capability} &  &   & 0.86 \\
                   \hline \hline
\textbf{        Bert Score} & Precision & Recall & F1 \\
                    \hline
                     & 0.63 & 0.67 & 0.65 \\
                            \hline \hline
    \end{tabular}
    \caption{Dermatology Evaluation scores}
    \label{tab:scores}
\end{table}

\begin{figure}
    \centering
    \includegraphics[width=0.8\linewidth]{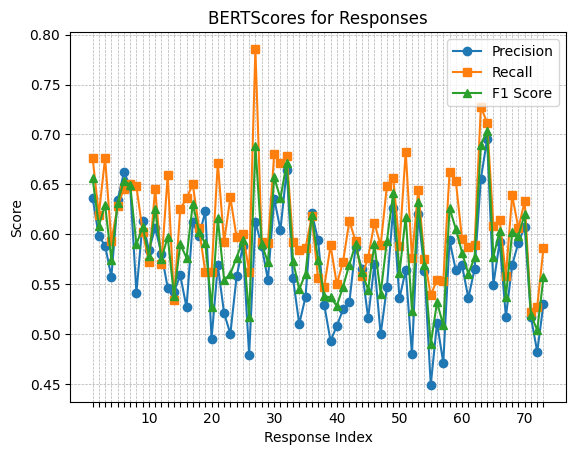}
    \caption{Bert Scores}
    \label{fig:bertScores}
\end{figure}

\begin{figure}
    \centering
    \includegraphics[width=1\linewidth]{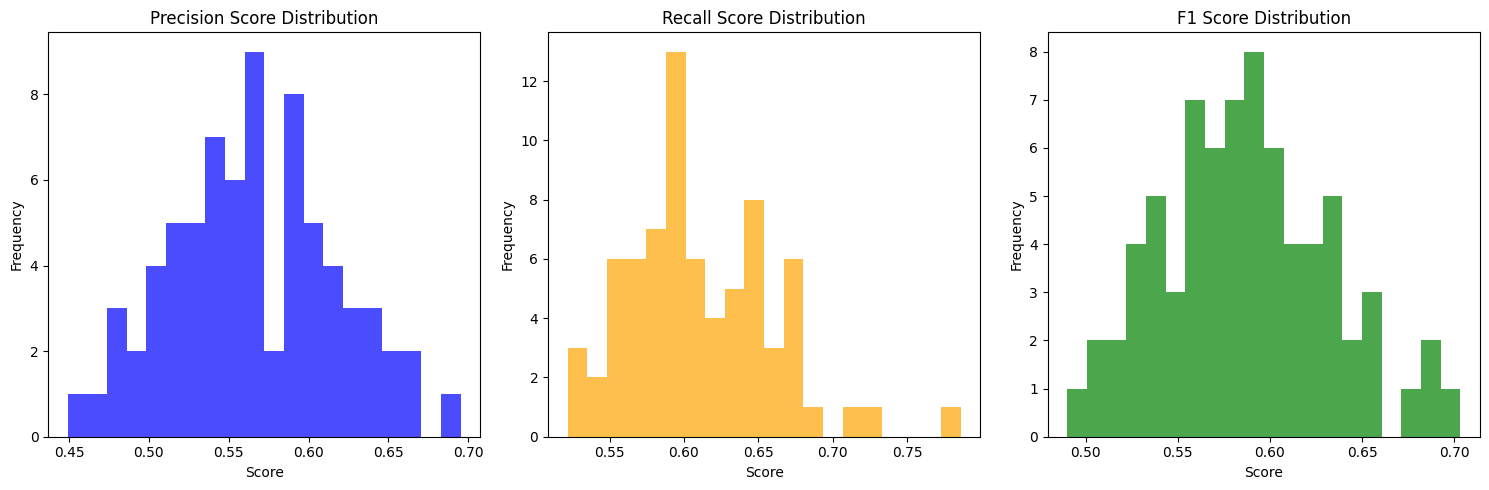}
    \caption{Bert Scores distribution}
    \label{fig:bertDistributions}
\end{figure}

The BERTScores, which also confirm the Textual similarity and NLI, suggest that the responses are somewhat aligned with the correct answers with room for improvement. This can be seen in table~\ref{tab:scores} and figures~\ref{fig:bertScores} and~\ref{fig:bertDistributions}. Precision in the context of BERTScore refers to the extent to which tokens in the generated response -hypothesis- are semantically similar to tokens in the premise. Recall measures how well the tokens in the correct answer are covered by semantically similar tokens in the response. On the other hand, the F1 Score is the harmonic mean of precision and recall. A moderate precision indicates that not all information in the response is relevant or accurately reflects the correct answer, while a higher recall suggests that a majority of the correct answer's content is addressed in the response, yet some aspects are missing or underrepresented. The F1 Score underscores the need for improvement in both precision and recall to achieve a higher degree of semantic similarity and accuracy. The BERTScores, alongside the textual -cosine- similarity and the NLI scores for context and diagnosis, show that there is good reasoning within the domain, where images are also analysed. This analysis can benefit the doctor's workflow and suggest pathways for further investigation, either via a more thorough clinical assessment or via external services, usually provided by laboratories or other experts. 

\subsubsection{Expert Review and Final Assessment}
As can be seen in table~\ref{tab:scores}, based on the expert review and following a thorough evaluation of the responses, the score validates our quality analysis based on textual similarity and NLI. For context, the score was equal to 4.38 (out of 5) or 0.87. For the entities, i.e. for the diagnosis reasoning, the score was 4.31 (out of 5) or  0.86 for an average score of 0.87. Based on our evaluation process, a final capability score is calculated and is equal to 0.86.

\section{Conclusion and future work}

In this paper, Dermacen Analytica was presented, which constitutes a novel assistive workflow integrating advanced technologies such as LLMs (namely, GPT-4), transformer-based vision models (namely, GPT-4V), and ML tools to enhance the process of evaluation of skin lesions and other conditions in dermatology. Through our comprehensive methodology, incorporating cross-model validation and advanced ML alongside NLP tools, we demonstrated significant improvements in diagnostic efficiency and accuracy. The system's ability to provide nuanced interpretation and detailed analysis of dermatological conditions may signify an advancement in tele-dermatology, as it allows leveraging both textual and visual data. Our evaluation, grounded on a thorough assessment pipeline and expert review, underscored the workflow's effectiveness, achieving very high (perhaps impressive) weighted scores in contextual understanding and diagnostic accuracy. For the evaluation, we employ NLP techniques to score similarity and NLI of correct and provided response and correct single or differential diagnosis and returned diagnosis. To accomplish that, we consider the default as the Premise and the returned as the Hypothesis, and we outline the process via a comprehensive algorithmic approach. 

Dermacen Analytica's deployment in enhancing dermatological care, especially in remote and underserved areas, underscores its potential to enhance the field with technological innovation. Indeed, the introduction of Dermacen Analytica into the tele-dermatology landscape may address some of the critical challenges in diagnostic accuracy and efficiency. By harnessing the capabilities of GPT-4V for reasoning and diagnosis, our approach not only streamlines the workflow for dermatologists, but it also enriches the decision-making process with AI-empowered insights. The methodology's emphasis on multi-model collaboration minimizes errors, such as hallucinations and misdiagnoses, thereby enhancing patient care outcomes. Moreover, the system's adaptability to different skin conditions and lesions, supported by evidence-based evaluation criteria, presents a scalable solution for future tele-dermatology applications. The potential of Dermacen Analytica to facilitate access to dermatological care, particularly in regions lacking specialized medical services, highlights the critical role of technology in bridging healthcare gaps.

However, the adoption of such advanced technological solutions also raises considerations around ethical use, data privacy, and the need for continuous validation against evolving dermatological knowledge. In our future work, we aim to address these challenges, ensuring that innovations like Dermacen Analytica not only advance medical practice but do so responsibly and ethically.

In conclusion, Dermacen Analytica exemplifies the transformative impact of integrating LLMs, vision transformers, and machine learning in healthcare. By elevating the standard of care in dermatology, this workflow paves the way for future advancements in telemedicine, promising a more efficient, accurate, and accessible healthcare landscape.

Future work will focus on refining AI models through continuous learning, expanding the range of conditions that can be accurately diagnosed, and integrating patient feedback to enhance the user experience. Furthermore, collaborations with dermatologists and healthcare institutions will be crucial in validating and adapting the workflow to meet clinical needs.


\label{comGap}

\section*{Acknowledgements}
This work has been partly supported by the University of Piraeus Research Center.

The authors would also like to extend their gratitude to the doctors of Dermacen SA~({\url{www.dermatologikokentro.gr/}}) for their technical assistance and guidance as per the various medical themes reported in this work. 

\section*{Data Availability}
Data sources utilized in this study have been cited and referenced. We are committed to contributing to the scientific community by open-sourcing both the engineered data and the code developed for this research, which will be made publicly available concurrent with the publication of our findings.

\section*{Abbreviations}
The following abbreviations are used in this paper:\\
\begin{tabular}{@{}ll}
ML & Machine learning \\
AI & Artificial Intelligence \\
IMA & Image Metadata Analysis \\
KG & Knowledge Graphs \\
XAI & Explainable Artificial Intelligence \\
NLP & Natural Language Processing\\
LLM & Large Language Model \\
MCQ & Multiple-choice Questionnaire \\
NER & Named Entity Recognition\\
ViT & Vision Transformer\\
CNN & Convolutional Neural Network\\
SKIN & Refers to conditions or diseases related to the skin\\
\end{tabular}

\bibliography{ref2}

\begin{thebibliography}{45}
\expandafter\ifx\csname natexlab\endcsname\relax\def\natexlab#1{#1}\fi
\providecommand{\url}[1]{\texttt{#1}}
\providecommand{\href}[2]{#2}
\providecommand{\path}[1]{#1}
\providecommand{\DOIprefix}{doi:}
\providecommand{\ArXivprefix}{arXiv:}
\providecommand{\URLprefix}{URL: }
\providecommand{\Pubmedprefix}{pmid:}
\providecommand{\doi}[1]{\href{http://dx.doi.org/#1}{\path{#1}}}
\providecommand{\Pubmed}[1]{\href{pmid:#1}{\path{#1}}}
\providecommand{\bibinfo}[2]{#2}
\ifx\xfnm\relax \def\xfnm[#1]{\unskip,\space#1}\fi
\bibitem[{Ahsan et~al.(2022)Ahsan, Luna \& Siddique}]{ahsan2022machine}
\bibinfo{author}{Ahsan, M.~M.}, \bibinfo{author}{Luna, S.~A.}, \& \bibinfo{author}{Siddique, Z.} (\bibinfo{year}{2022}).
\newblock \bibinfo{title}{Machine-learning-based disease diagnosis: A comprehensive review}.
\newblock In {\it \bibinfo{booktitle}{Healthcare}\/} (p. \bibinfo{pages}{541}).
\newblock \bibinfo{organization}{MDPI} volume~\bibinfo{volume}{10}.
\bibitem[{American Lung association()}]{dermnetnz}
American Lung association ().
\newblock \bibinfo{title}{American lung association}.
\newblock \bibinfo{howpublished}{\url{https://dermnetnz.org/about}}.
\newblock \bibinfo{note}{Accessed: 2023-05-29}.
\bibitem[{Anand et~al.(2023)Anand, Gupta, Koundal \& Singh}]{anand2023fusion}
\bibinfo{author}{Anand, V.}, \bibinfo{author}{Gupta, S.}, \bibinfo{author}{Koundal, D.}, \& \bibinfo{author}{Singh, K.} (\bibinfo{year}{2023}).
\newblock \bibinfo{title}{Fusion of u-net and cnn model for segmentation and classification of skin lesion from dermoscopy images}.
\newblock {\it \bibinfo{journal}{Expert Systems with Applications}\/},  {\it \bibinfo{volume}{213}\/}, \bibinfo{pages}{119230}.
\bibitem[{Atito et~al.(2021)Atito, Awais \& Kittler}]{atito2021sit}
\bibinfo{author}{Atito, S.}, \bibinfo{author}{Awais, M.}, \& \bibinfo{author}{Kittler, J.} (\bibinfo{year}{2021}).
\newblock \bibinfo{title}{Sit: Self-supervised vision transformer}.
\newblock {\it \bibinfo{journal}{arXiv preprint arXiv:2104.03602}\/}, .
\bibitem[{Bradski et~al.(2000)Bradski, Kaehler et~al.}]{bradski2000opencv}
\bibinfo{author}{Bradski, G.}, \bibinfo{author}{Kaehler, A.} et~al. (\bibinfo{year}{2000}).
\newblock \bibinfo{title}{Opencv}.
\newblock {\it \bibinfo{journal}{Dr. Dobb’s journal of software tools}\/},  {\it \bibinfo{volume}{3}\/}.
\bibitem[{Chan et~al.(2020)Chan, Reddy, Myers, Thibodeaux, Brownstone \& Liao}]{chan2020machine}
\bibinfo{author}{Chan, S.}, \bibinfo{author}{Reddy, V.}, \bibinfo{author}{Myers, B.}, \bibinfo{author}{Thibodeaux, Q.}, \bibinfo{author}{Brownstone, N.}, \& \bibinfo{author}{Liao, W.} (\bibinfo{year}{2020}).
\newblock \bibinfo{title}{Machine learning in dermatology: current applications, opportunities, and limitations}.
\newblock {\it \bibinfo{journal}{Dermatology and therapy}\/},  {\it \bibinfo{volume}{10}\/}, \bibinfo{pages}{365--386}.
\bibitem[{Devlin et~al.(2018)Devlin, Chang, Lee \& Toutanova}]{devlin2018bert}
\bibinfo{author}{Devlin, J.}, \bibinfo{author}{Chang, M.-W.}, \bibinfo{author}{Lee, K.}, \& \bibinfo{author}{Toutanova, K.} (\bibinfo{year}{2018}).
\newblock \bibinfo{title}{Bert: Pre-training of deep bidirectional transformers for language understanding}.
\newblock {\it \bibinfo{journal}{arXiv preprint arXiv:1810.04805}\/}, .
\bibitem[{Du et~al.(2020)Du, Emam \& Gniadecki}]{du2020review}
\bibinfo{author}{Du, A.~X.}, \bibinfo{author}{Emam, S.}, \& \bibinfo{author}{Gniadecki, R.} (\bibinfo{year}{2020}).
\newblock \bibinfo{title}{Review of machine learning in predicting dermatological outcomes}.
\newblock {\it \bibinfo{journal}{Frontiers in Medicine}\/},  {\it \bibinfo{volume}{7}\/}, \bibinfo{pages}{266}.
\bibitem[{Duarte et~al.(2021)Duarte, Sousa-Pinto, Azevedo, Barros, Puig, Malvehy, Haneke \& Correia}]{duarte2021clinical}
\bibinfo{author}{Duarte, A.~F.}, \bibinfo{author}{Sousa-Pinto, B.}, \bibinfo{author}{Azevedo, L.~F.}, \bibinfo{author}{Barros, A.~M.}, \bibinfo{author}{Puig, S.}, \bibinfo{author}{Malvehy, J.}, \bibinfo{author}{Haneke, E.}, \& \bibinfo{author}{Correia, O.} (\bibinfo{year}{2021}).
\newblock \bibinfo{title}{Clinical abcde rule for early melanoma detection}.
\newblock {\it \bibinfo{journal}{European Journal of Dermatology}\/},  {\it \bibinfo{volume}{31}\/}, \bibinfo{pages}{771--778}.
\bibitem[{Fan et~al.(2017)Fan, Xie, Li, Jiang \& Liu}]{fan2017automatic}
\bibinfo{author}{Fan, H.}, \bibinfo{author}{Xie, F.}, \bibinfo{author}{Li, Y.}, \bibinfo{author}{Jiang, Z.}, \& \bibinfo{author}{Liu, J.} (\bibinfo{year}{2017}).
\newblock \bibinfo{title}{Automatic segmentation of dermoscopy images using saliency combined with otsu threshold}.
\newblock {\it \bibinfo{journal}{Computers in biology and medicine}\/},  {\it \bibinfo{volume}{85}\/}, \bibinfo{pages}{75--85}.
\bibitem[{Ferris \& O'Flynn(2015)}]{ferris2015assessment}
\bibinfo{author}{Ferris, H.}, \& \bibinfo{author}{O'Flynn, D.} (\bibinfo{year}{2015}).
\newblock \bibinfo{title}{Assessment in medical education; what are we trying to achieve?.}
\newblock {\it \bibinfo{journal}{International Journal of Higher Education}\/},  {\it \bibinfo{volume}{4}\/}, \bibinfo{pages}{139--144}.
\bibitem[{Gachon et~al.(2005)Gachon, Beaulieu, Sei, Gouvernet, Claudel, Lemaitre, Richard \& Grob}]{gachon2005first}
\bibinfo{author}{Gachon, J.}, \bibinfo{author}{Beaulieu, P.}, \bibinfo{author}{Sei, J.~F.}, \bibinfo{author}{Gouvernet, J.}, \bibinfo{author}{Claudel, J.~P.}, \bibinfo{author}{Lemaitre, M.}, \bibinfo{author}{Richard, M.~A.}, \& \bibinfo{author}{Grob, J.~J.} (\bibinfo{year}{2005}).
\newblock \bibinfo{title}{First prospective study of the recognition process of melanoma in dermatological practice}.
\newblock {\it \bibinfo{journal}{Archives of dermatology}\/},  {\it \bibinfo{volume}{141}\/}, \bibinfo{pages}{434--438}.
\bibitem[{Hahnel et~al.(2017)Hahnel, Lichterfeld, Blume-Peytavi \& Kottner}]{hahnel2017epidemiology}
\bibinfo{author}{Hahnel, E.}, \bibinfo{author}{Lichterfeld, A.}, \bibinfo{author}{Blume-Peytavi, U.}, \& \bibinfo{author}{Kottner, J.} (\bibinfo{year}{2017}).
\newblock \bibinfo{title}{The epidemiology of skin conditions in the aged: a systematic review}.
\newblock {\it \bibinfo{journal}{Journal of tissue viability}\/},  {\it \bibinfo{volume}{26}\/}, \bibinfo{pages}{20--28}.
\bibitem[{Hay et~al.(2014)Hay, Johns, Williams, Bolliger, Dellavalle, Margolis, Marks, Naldi, Weinstock, Wulf et~al.}]{hay2014global}
\bibinfo{author}{Hay, R.~J.}, \bibinfo{author}{Johns, N.~E.}, \bibinfo{author}{Williams, H.~C.}, \bibinfo{author}{Bolliger, I.~W.}, \bibinfo{author}{Dellavalle, R.~P.}, \bibinfo{author}{Margolis, D.~J.}, \bibinfo{author}{Marks, R.}, \bibinfo{author}{Naldi, L.}, \bibinfo{author}{Weinstock, M.~A.}, \bibinfo{author}{Wulf, S.~K.} et~al. (\bibinfo{year}{2014}).
\newblock \bibinfo{title}{The global burden of skin disease in 2010: an analysis of the prevalence and impact of skin conditions}.
\newblock {\it \bibinfo{journal}{Journal of Investigative Dermatology}\/},  {\it \bibinfo{volume}{134}\/}, \bibinfo{pages}{1527--1534}.
\bibitem[{Hogarty et~al.(2020)Hogarty, Su, Phan, Attia, Hossny, Nahavandi, Lenane, Moloney \& Yazdabadi}]{hogarty2020artificial}
\bibinfo{author}{Hogarty, D.~T.}, \bibinfo{author}{Su, J.~C.}, \bibinfo{author}{Phan, K.}, \bibinfo{author}{Attia, M.}, \bibinfo{author}{Hossny, M.}, \bibinfo{author}{Nahavandi, S.}, \bibinfo{author}{Lenane, P.}, \bibinfo{author}{Moloney, F.~J.}, \& \bibinfo{author}{Yazdabadi, A.} (\bibinfo{year}{2020}).
\newblock \bibinfo{title}{Artificial intelligence in dermatology—where we are and the way to the future: a review}.
\newblock {\it \bibinfo{journal}{American journal of clinical dermatology}\/},  {\it \bibinfo{volume}{21}\/}, \bibinfo{pages}{41--47}.
\bibitem[{Hong et~al.(2020)Hong, Tchoua, Chard \& Foster}]{hong2020sciner}
\bibinfo{author}{Hong, Z.}, \bibinfo{author}{Tchoua, R.}, \bibinfo{author}{Chard, K.}, \& \bibinfo{author}{Foster, I.} (\bibinfo{year}{2020}).
\newblock \bibinfo{title}{Sciner: extracting named entities from scientific literature}.
\newblock In {\it \bibinfo{booktitle}{Computational Science--ICCS 2020: 20th International Conference, Amsterdam, The Netherlands, June 3--5, 2020, Proceedings, Part II 20}\/} (pp. \bibinfo{pages}{308--321}).
\newblock \bibinfo{organization}{Springer}.
\bibitem[{Iranpoor et~al.(2020)Iranpoor, Mahboob, Shahbandegan \& Baniasadi}]{iranpoor2020skin}
\bibinfo{author}{Iranpoor, R.}, \bibinfo{author}{Mahboob, A.~S.}, \bibinfo{author}{Shahbandegan, S.}, \& \bibinfo{author}{Baniasadi, N.} (\bibinfo{year}{2020}).
\newblock \bibinfo{title}{Skin lesion segmentation using convolutional neural networks with improved u-net architecture}.
\newblock In {\it \bibinfo{booktitle}{2020 6th Iranian Conference on Signal Processing and Intelligent Systems (ICSPIS)}\/} (pp. \bibinfo{pages}{1--5}).
\newblock \bibinfo{organization}{IEEE}.
\bibitem[{Johnson et~al.(2021)Johnson, Wei, Weeraratne, Frisse, Misulis, Rhee, Zhao \& Snowdon}]{johnson2021precision}
\bibinfo{author}{Johnson, K.~B.}, \bibinfo{author}{Wei, W.-Q.}, \bibinfo{author}{Weeraratne, D.}, \bibinfo{author}{Frisse, M.~E.}, \bibinfo{author}{Misulis, K.}, \bibinfo{author}{Rhee, K.}, \bibinfo{author}{Zhao, J.}, \& \bibinfo{author}{Snowdon, J.~L.} (\bibinfo{year}{2021}).
\newblock \bibinfo{title}{Precision medicine, ai, and the future of personalized health care}.
\newblock {\it \bibinfo{journal}{Clinical and translational science}\/},  {\it \bibinfo{volume}{14}\/}, \bibinfo{pages}{86--93}.
\bibitem[{Kerr et~al.(2010)Kerr, Tidman, Walker, Aldridge \& Benton}]{kerr2010profile}
\bibinfo{author}{Kerr, O.}, \bibinfo{author}{Tidman, M.}, \bibinfo{author}{Walker, J.}, \bibinfo{author}{Aldridge, R.}, \& \bibinfo{author}{Benton, E.} (\bibinfo{year}{2010}).
\newblock \bibinfo{title}{The profile of dermatological problems in primary care}.
\newblock {\it \bibinfo{journal}{Clinical and experimental dermatology}\/},  {\it \bibinfo{volume}{35}\/}, \bibinfo{pages}{380--383}.
\bibitem[{Khan et~al.(2022)Khan, Naseer, Hayat, Zamir, Khan \& Shah}]{khan2022transformers}
\bibinfo{author}{Khan, S.}, \bibinfo{author}{Naseer, M.}, \bibinfo{author}{Hayat, M.}, \bibinfo{author}{Zamir, S.~W.}, \bibinfo{author}{Khan, F.~S.}, \& \bibinfo{author}{Shah, M.} (\bibinfo{year}{2022}).
\newblock \bibinfo{title}{Transformers in vision: A survey}.
\newblock {\it \bibinfo{journal}{ACM computing surveys (CSUR)}\/},  {\it \bibinfo{volume}{54}\/}, \bibinfo{pages}{1--41}.
\bibitem[{Labrak et~al.(2024)Labrak, Bazoge, Morin, Gourraud, Rouvier \& Dufour}]{labrak2024biomistral}
\bibinfo{author}{Labrak, Y.}, \bibinfo{author}{Bazoge, A.}, \bibinfo{author}{Morin, E.}, \bibinfo{author}{Gourraud, P.-A.}, \bibinfo{author}{Rouvier, M.}, \& \bibinfo{author}{Dufour, R.} (\bibinfo{year}{2024}).
\newblock \bibinfo{title}{Biomistral: A collection of open-source pretrained large language models for medical domains}.
\newblock {\it \bibinfo{journal}{arXiv preprint arXiv:2402.10373}\/}, .
\bibitem[{Liopyris et~al.(2022)Liopyris, Gregoriou, Dias \& Stratigos}]{liopyris2022artificial}
\bibinfo{author}{Liopyris, K.}, \bibinfo{author}{Gregoriou, S.}, \bibinfo{author}{Dias, J.}, \& \bibinfo{author}{Stratigos, A.~J.} (\bibinfo{year}{2022}).
\newblock \bibinfo{title}{Artificial intelligence in dermatology: challenges and perspectives}.
\newblock {\it \bibinfo{journal}{Dermatology and Therapy}\/},  {\it \bibinfo{volume}{12}\/}, \bibinfo{pages}{2637--2651}.
\bibitem[{Liu et~al.(2019)Liu, Ott, Goyal, Du, Joshi, Chen, Levy, Lewis, Zettlemoyer \& Stoyanov}]{liu2019roberta}
\bibinfo{author}{Liu, Y.}, \bibinfo{author}{Ott, M.}, \bibinfo{author}{Goyal, N.}, \bibinfo{author}{Du, J.}, \bibinfo{author}{Joshi, M.}, \bibinfo{author}{Chen, D.}, \bibinfo{author}{Levy, O.}, \bibinfo{author}{Lewis, M.}, \bibinfo{author}{Zettlemoyer, L.}, \& \bibinfo{author}{Stoyanov, V.} (\bibinfo{year}{2019}).
\newblock \bibinfo{title}{Roberta: A robustly optimized bert pretraining approach}.
\newblock {\it \bibinfo{journal}{arXiv preprint arXiv:1907.11692}\/}, .
\bibitem[{Machine Learning for PyTorch, TensorFlow, and JAX()}]{hugging}
Machine Learning for PyTorch, TensorFlow, and JAX ().
\newblock \bibinfo{title}{State-of--the--art}.
\newblock \bibinfo{howpublished}{\url{https://huggingface.co/docs/transformers/en/index}}.
\newblock \bibinfo{note}{Accessed: 2024-3-15}.
\bibitem[{Majumder et~al.(2019)Majumder, Kumar, Krishnamurthy, Ojeh, Adams \& Sa}]{majumder2019evaluative}
\bibinfo{author}{Majumder, M. A.~A.}, \bibinfo{author}{Kumar, A.}, \bibinfo{author}{Krishnamurthy, K.}, \bibinfo{author}{Ojeh, N.}, \bibinfo{author}{Adams, O.~P.}, \& \bibinfo{author}{Sa, B.} (\bibinfo{year}{2019}).
\newblock \bibinfo{title}{An evaluative study of objective structured clinical examination (osce): students and examiners perspectives}.
\newblock {\it \bibinfo{journal}{Advances in medical education and practice}\/},  (pp. \bibinfo{pages}{387--397}).
\bibitem[{Marwaha et~al.(2019)Marwaha, Fevrier, Alexeeff, Crowley, Haiman, Pham, Tuerk, Wukda, Hartmann \& Herrinton}]{marwaha2019comparative}
\bibinfo{author}{Marwaha, S.~S.}, \bibinfo{author}{Fevrier, H.}, \bibinfo{author}{Alexeeff, S.}, \bibinfo{author}{Crowley, E.}, \bibinfo{author}{Haiman, M.}, \bibinfo{author}{Pham, N.}, \bibinfo{author}{Tuerk, M.~J.}, \bibinfo{author}{Wukda, D.}, \bibinfo{author}{Hartmann, M.}, \& \bibinfo{author}{Herrinton, L.~J.} (\bibinfo{year}{2019}).
\newblock \bibinfo{title}{Comparative effectiveness study of face-to-face and teledermatology workflows for diagnosing skin cancer}.
\newblock {\it \bibinfo{journal}{Journal of the American Academy of Dermatology}\/},  {\it \bibinfo{volume}{81}\/}, \bibinfo{pages}{1099--1106}.
\bibitem[{Nadeau \& Sekine(2007)}]{nadeau2007survey}
\bibinfo{author}{Nadeau, D.}, \& \bibinfo{author}{Sekine, S.} (\bibinfo{year}{2007}).
\newblock \bibinfo{title}{A survey of named entity recognition and classification}.
\newblock {\it \bibinfo{journal}{Lingvisticae Investigationes}\/},  {\it \bibinfo{volume}{30}\/}, \bibinfo{pages}{3--26}.
\bibitem[{Norcini \& McKinley(2007)}]{norcini2007assessment}
\bibinfo{author}{Norcini, J.~J.}, \& \bibinfo{author}{McKinley, D.~W.} (\bibinfo{year}{2007}).
\newblock \bibinfo{title}{Assessment methods in medical education}.
\newblock {\it \bibinfo{journal}{Teaching and teacher education}\/},  {\it \bibinfo{volume}{23}\/}, \bibinfo{pages}{239--250}.
\bibitem[{Panagoulias et~al.(2023{\natexlab{a}})Panagoulias, Palamidas, Virvou \& Tsihrintzis}]{panagoulias2023IISALlms}
\bibinfo{author}{Panagoulias, D.}, \bibinfo{author}{Palamidas, F.}, \bibinfo{author}{Virvou, M.}, \& \bibinfo{author}{Tsihrintzis, G.~A.} (\bibinfo{year}{2023}{\natexlab{a}}).
\newblock \bibinfo{title}{Evaluating the potential of llms and chatgpt on medical diagnosis and treatment}.
\newblock In {\it \bibinfo{booktitle}{14th IEEE International Conference on Information, Intelligence, Systems, and Applications (IISA2023), Volos, Greece}\/}.
\bibitem[{Panagoulias et~al.(2023{\natexlab{b}})Panagoulias, Virvou \& Tsihrintzis}]{ruleAugm}
\bibinfo{author}{Panagoulias, D.~P.}, \bibinfo{author}{Virvou, M.}, \& \bibinfo{author}{Tsihrintzis, G.~A.} (\bibinfo{year}{2023}{\natexlab{b}}).
\newblock \bibinfo{title}{Rule-augmented artificial intelligence-empowered systems for medical diagnosis using large language models}.
\newblock In {\it \bibinfo{booktitle}{2023 IEEE 35th International Conference on Tools with Artificial Intelligence (ICTAI)}\/}.
\bibitem[{Panagoulias et~al.(2024{\natexlab{a}})Panagoulias, Virvou \& Tsihrintzis}]{panagoulias2024evaluating}
\bibinfo{author}{Panagoulias, D.~P.}, \bibinfo{author}{Virvou, M.}, \& \bibinfo{author}{Tsihrintzis, G.~A.} (\bibinfo{year}{2024}{\natexlab{a}}).
\newblock \bibinfo{title}{Evaluating llm -- generated multimodal diagnosis from medical images and symptom analysis}.
\newblock \href{http://arxiv.org/abs/2402.01730}{\tt arXiv:2402.01730}.
\bibitem[{Panagoulias et~al.(2024{\natexlab{b}})Panagoulias, Virvou \& Tsihrintzis}]{PANAGOULIAS2024123375}
\bibinfo{author}{Panagoulias, D.~P.}, \bibinfo{author}{Virvou, M.}, \& \bibinfo{author}{Tsihrintzis, G.~A.} (\bibinfo{year}{2024}{\natexlab{b}}).
\newblock \bibinfo{title}{A novel framework for artificial intelligence explainability via the technology acceptance model and rapid estimate of adult literacy in medicine using machine learning}.
\newblock {\it \bibinfo{journal}{Expert Systems with Applications}\/},  {\it \bibinfo{volume}{248}\/}, \bibinfo{pages}{123375}. \URLprefix \url{https://www.sciencedirect.com/science/article/pii/S0957417424002409}. \DOIprefix\doi{https://doi.org/10.1016/j.eswa.2024.123375}.
\bibitem[{Reimers \& Gurevych(2019)}]{reimers-2019-sentence-bert}
\bibinfo{author}{Reimers, N.}, \& \bibinfo{author}{Gurevych, I.} (\bibinfo{year}{2019}).
\newblock \bibinfo{title}{Sentence-bert: Sentence embeddings using siamese bert-networks}.
\newblock In {\it \bibinfo{booktitle}{Proceedings of the 2019 Conference on Empirical Methods in Natural Language Processing}\/}.
\newblock \bibinfo{publisher}{Association for Computational Linguistics}.
\newblock \URLprefix \url{https://arxiv.org/abs/1908.10084}.
\bibitem[{Ring et~al.(2021)Ring, Cox \& Lee}]{RING2021635}
\bibinfo{author}{Ring, C.}, \bibinfo{author}{Cox, N.}, \& \bibinfo{author}{Lee, J.~B.} (\bibinfo{year}{2021}).
\newblock \bibinfo{title}{Dermatoscopy}.
\newblock {\it \bibinfo{journal}{Clinics in Dermatology}\/},  {\it \bibinfo{volume}{39}\/}, \bibinfo{pages}{635--642}. \URLprefix \url{https://www.sciencedirect.com/science/article/pii/S0738081X21000547}. \DOIprefix\doi{https://doi.org/10.1016/j.clindermatol.2021.03.009}.
\newblock \bibinfo{note}{Diagnostic and therapeutic instrumentation in dermatology}.
\bibitem[{Setiawan(2020)}]{setiawan2020image}
\bibinfo{author}{Setiawan, A.~W.} (\bibinfo{year}{2020}).
\newblock \bibinfo{title}{Image segmentation metrics in skin lesion: accuracy, sensitivity, specificity, dice coefficient, jaccard index, and matthews correlation coefficient}.
\newblock In {\it \bibinfo{booktitle}{2020 International Conference on Computer Engineering, Network, and Intelligent Multimedia (CENIM)}\/} (pp. \bibinfo{pages}{97--102}).
\newblock \bibinfo{organization}{IEEE}.
\bibitem[{System Card()}]{2023GPT4VisionSC}
System Card (\bibinfo{year}{2023}).
\newblock \bibinfo{title}{Gpt-4v(ision)}.
\bibitem[{Vanderbilt et~al.(2013)Vanderbilt, Feldman \& Wood}]{vanderbilt2013assessment}
\bibinfo{author}{Vanderbilt, A.}, \bibinfo{author}{Feldman, M.}, \& \bibinfo{author}{Wood, I.} (\bibinfo{year}{2013}).
\newblock \bibinfo{title}{Assessment in undergraduate medical education: a review of course exams}.
\newblock {\it \bibinfo{journal}{Medical education online}\/},  {\it \bibinfo{volume}{18}\/}, \bibinfo{pages}{20438}.
\bibitem[{Vaswani et~al.(2017)Vaswani, Shazeer, Parmar, Uszkoreit, Jones, Gomez, Kaiser \& Polosukhin}]{vaswani2017attention}
\bibinfo{author}{Vaswani, A.}, \bibinfo{author}{Shazeer, N.}, \bibinfo{author}{Parmar, N.}, \bibinfo{author}{Uszkoreit, J.}, \bibinfo{author}{Jones, L.}, \bibinfo{author}{Gomez, A.~N.}, \bibinfo{author}{Kaiser, {\L}.}, \& \bibinfo{author}{Polosukhin, I.} (\bibinfo{year}{2017}).
\newblock \bibinfo{title}{Attention is all you need}.
\newblock In {\it \bibinfo{booktitle}{Advances in Neural Information Processing Systems 30: Annual Conference on Neural Information Processing Systems 2017, December 4-9, 2017, Long Beach, CA, {USA}}\/} (pp. \bibinfo{pages}{5998--6008}).
\newblock volume~\bibinfo{volume}{30}.
\bibitem[{Virvou(2023)}]{virvouartificial}
\bibinfo{author}{Virvou, M.} (\bibinfo{year}{2023}).
\newblock \bibinfo{title}{Artificial intelligence and user experience in reciprocity: Contributions and state of the art}.
\newblock {\it \bibinfo{journal}{Intelligent Decision Technologies}\/},  {\it \bibinfo{volume}{17}\/}, \bibinfo{pages}{73--125}.
\bibitem[{Wang et~al.(2022)Wang, Lei, Cui, Zhang, Meng \& Nandi}]{wang2022medical}
\bibinfo{author}{Wang, R.}, \bibinfo{author}{Lei, T.}, \bibinfo{author}{Cui, R.}, \bibinfo{author}{Zhang, B.}, \bibinfo{author}{Meng, H.}, \& \bibinfo{author}{Nandi, A.~K.} (\bibinfo{year}{2022}).
\newblock \bibinfo{title}{Medical image segmentation using deep learning: A survey}.
\newblock {\it \bibinfo{journal}{IET Image Processing}\/},  {\it \bibinfo{volume}{16}\/}, \bibinfo{pages}{1243--1267}.
\bibitem[{Williams et~al.(2018)Williams, Nangia \& Bowman}]{N18-1101}
\bibinfo{author}{Williams, A.}, \bibinfo{author}{Nangia, N.}, \& \bibinfo{author}{Bowman, S.} (\bibinfo{year}{2018}).
\newblock \bibinfo{title}{A broad-coverage challenge corpus for sentence understanding through inference}.
\newblock In {\it \bibinfo{booktitle}{Proceedings of the 2018 Conference of the North American Chapter of the Association for Computational Linguistics: Human Language Technologies, Volume 1 (Long Papers)}\/} (pp. \bibinfo{pages}{1112--1122}).
\newblock \bibinfo{publisher}{Association for Computational Linguistics}.
\newblock \URLprefix \url{http://aclweb.org/anthology/N18-1101}.
\bibitem[{Yuan et~al.(2021)Yuan, Chen, Wang, Yu, Shi, Jiang, Tay, Feng \& Yan}]{yuan2021tokens}
\bibinfo{author}{Yuan, L.}, \bibinfo{author}{Chen, Y.}, \bibinfo{author}{Wang, T.}, \bibinfo{author}{Yu, W.}, \bibinfo{author}{Shi, Y.}, \bibinfo{author}{Jiang, Z.-H.}, \bibinfo{author}{Tay, F.~E.}, \bibinfo{author}{Feng, J.}, \& \bibinfo{author}{Yan, S.} (\bibinfo{year}{2021}).
\newblock \bibinfo{title}{Tokens-to-token vit: Training vision transformers from scratch on imagenet}.
\newblock In {\it \bibinfo{booktitle}{Proceedings of the IEEE/CVF international conference on computer vision}\/} (pp. \bibinfo{pages}{558--567}).
\bibitem[{Zhai et~al.(2022)Zhai, Kolesnikov, Houlsby \& Beyer}]{zhai2022scaling}
\bibinfo{author}{Zhai, X.}, \bibinfo{author}{Kolesnikov, A.}, \bibinfo{author}{Houlsby, N.}, \& \bibinfo{author}{Beyer, L.} (\bibinfo{year}{2022}).
\newblock \bibinfo{title}{Scaling vision transformers}.
\newblock In {\it \bibinfo{booktitle}{Proceedings of the IEEE/CVF Conference on Computer Vision and Pattern Recognition}\/} (pp. \bibinfo{pages}{12104--12113}).
\bibitem[{Zhang et~al.(2018)Zhang, Zhao \& Bai}]{zhang2018object}
\bibinfo{author}{Zhang, S.}, \bibinfo{author}{Zhao, Y.}, \& \bibinfo{author}{Bai, P.} (\bibinfo{year}{2018}).
\newblock \bibinfo{title}{Object localization improved grabcut for lung parenchyma segmentation}.
\newblock {\it \bibinfo{journal}{Procedia computer science}\/},  {\it \bibinfo{volume}{131}\/}, \bibinfo{pages}{1311--1317}.
\bibitem[{Zhang* et~al.(2020)Zhang*, Kishore*, Wu*, Weinberger \& Artzi}]{bert-score}
\bibinfo{author}{Zhang*, T.}, \bibinfo{author}{Kishore*, V.}, \bibinfo{author}{Wu*, F.}, \bibinfo{author}{Weinberger, K.~Q.}, \& \bibinfo{author}{Artzi, Y.} (\bibinfo{year}{2020}).
\newblock \bibinfo{title}{Bertscore: Evaluating text generation with bert}.
\newblock In {\it \bibinfo{booktitle}{International Conference on Learning Representations}\/}.
\newblock \URLprefix \url{https://openreview.net/forum?id=SkeHuCVFDr}.

\end{thebibliography}

\end{document}